\documentclass[11pt]{article}

\usepackage[final]{acl}

\usepackage{times}
\usepackage{latexsym}

\usepackage[T1]{fontenc}

\usepackage[utf8]{inputenc}

\usepackage{microtype}

\usepackage{inconsolata}

\usepackage{graphicx}

\usepackage{url}            
\usepackage{booktabs}       
\usepackage{amsfonts}       
\usepackage{nicefrac}       
\usepackage{microtype}      
\usepackage{xcolor}         
\usepackage{setspace}
\usepackage{enumitem}
\usepackage{wrapfig}
\usepackage{caption}
\usepackage{amsmath} 
\usepackage{algorithm}
\usepackage{algpseudocode}
\usepackage{mathtools}
\usepackage{makecell}
\usepackage{subcaption}
\usepackage{multirow}
\usepackage{CJK}
\usepackage{soul}
\usepackage{etoolbox}
\usepackage[svgnames,table]{xcolor}
\usepackage{multirow}
\usepackage{array} 
\usepackage{caption}
\usepackage{tabularx}
\usepackage[most]{tcolorbox}
\usepackage{lipsum}

\usepackage{hyperref}
\usepackage{cleveref}
\usepackage{xspace}
\usepackage{listings}
\usepackage{xcolor}

\definecolor{auditkw}{rgb}{0.18,0.43,0.71}     
\definecolor{audittype}{rgb}{0.30,0.55,0.30}   
\definecolor{auditcom}{rgb}{0.45,0.45,0.45}    
\definecolor{auditstr}{rgb}{0.65,0.30,0.30}    
\definecolor{auditbg}{rgb}{0.980,0.988,0.992}  

\lstdefinestyle{auditpython}{
    language=Python,
    basicstyle=\ttfamily\footnotesize,
    keywordstyle=\color{auditkw}\bfseries,
    commentstyle=\color{auditcom}\itshape,
    stringstyle=\color{auditstr},
    emphstyle=\color{audittype}\bfseries,
    emph={str,float,list,BaseModel,Factor},
    backgroundcolor=\color{auditbg},
    showstringspaces=false,
    columns=fullflexible,
    keepspaces=true,
    breaklines=true,
    breakatwhitespace=true,
    breakindent=10pt,
    frame=single,                       
    rulecolor=\color{gray!50},          
    framerule=0.4pt,                    
    framesep=4pt,                       
    aboveskip=4pt, belowskip=4pt,
    xleftmargin=4pt, xrightmargin=4pt,
}

\definecolor{mypurple}{HTML}{8E8BFE}
\definecolor{mygreen}{HTML}{00B7EB}

\newcommand{\ifcomments}{\iftrue}

\newcommand{\method}{\textsc{AuditForecast}\xspace}

\title{From Narrative to Auditable Forecasts:\\A Structured Scaffold for Agentic Forecasting}

\author{Yuanpu Cao, Yongkang Du, Yurui Chang, Lu Lin, Jinghui Chen \\
The Pennsylvania State University\\
\texttt{\{ymc5533,ybd5136,yuruic,lulin,jzc5917\}@psu.edu} 
}

\begin{document}
\maketitle

\begin{abstract}
LLM agents are increasingly used for live forecasting, where they retrieve up-to-date information and produce estimates for unresolved future events. However, current agentic forecasting often relies on implicit narrative aggregation: agents collect evidence, discuss it in prose, and often assign a probability without an explicit update path from evidence to forecast. This limits both forecasting accuracy and auditability. We propose \textsc{AuditForecast}, an agentic scaffold for structured probabilistic forecasting. \textsc{AuditForecast} first anchors the forecast with a suitable quantitative baseline model, uses model-guided data retrieval to derive a base probability, and then applies situational factor updates outside the model's scope through mechanical aggregation in odds space. This turns forecasting from a prose-based judgment into a structured process with explicit intermediate objects. Across multiple live forecasting benchmarks, \textsc{AuditForecast} improves forecasting accuracy and calibration relative to strong agentic baselines, surpasses market-implied references in several settings, and outperforms substantially more expensive deep-research agents while remaining Pareto-dominant in the cost--accuracy tradeoff. Beyond performance gains, \textsc{AuditForecast} produces an auditable forecasting report that makes forecast construction explicit and supports systematic post hoc analysis.
\end{abstract}

\section{Introduction}
Large language model (LLM) agents~\citep{react,agent-survey} augment foundation models with action modules for interacting with external environments \citep{toolformer}, enabling them to perform web search, call APIs, and use task-specific tools to tackle more demanding real-world tasks \citep{webexplorer,sweagent}. In this context, forecasting future events has emerged as an increasingly prominent application for LLM agents \citep{futurex}, since such predictions are closely tied to decision-making in domains such as economics \citep{finance}, politics \citep{policy}, and business \citep{business}, with broad consequences for markets and public life. 

\begin{figure}[t]
\begin{center}
\centerline{\includegraphics[width=1.0\linewidth]{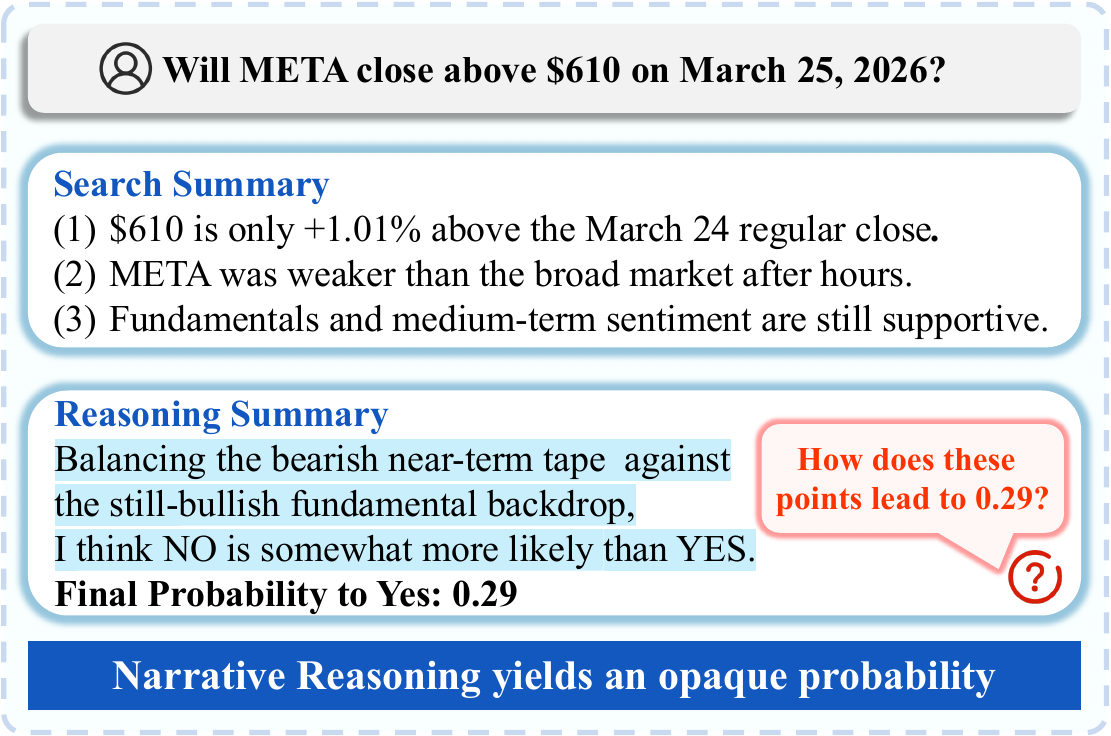}}
\vskip -0.05in
\caption{A narrative forecast from GPT-5.4 under Think \& Search agent framework: the agent retrieves evidence, balances it narratively, and assigns a probability without an explicit derivation.}
\label{fig:baseline_case}
\end{center}
\vskip -0.25in
\end{figure}
At the same time, forecasting is a challenging task for LLM agents, requiring high-quality information gathering, evidence analysis, and probabilistic reasoning under uncertainty. To evaluate these capabilities, recent work has introduced contamination-free live benchmarks \citep{prophet,futurex} by extracting forecasting questions from real-time public prediction platforms and evaluating advanced LLM agents on them. In particular, FutureX \citep{futurex} shows that both Think\&Search agents and deep-research agents can improve forecasting accuracy over base LLMs. Specifically, Think\&Search agents are built on the ReAct framework \citep{react}, executing iterative Reason--Act--Observe cycles to incrementally acquire new information through web search and refine their forecasts. Deep-research agents \citep{deepresearch} extend this scaffold with longer-horizon planning and specialized subagents, enabling more hierarchical information gathering and reasoning.

However, even though these agents actively search for and accumulate substantial contextual information and evidence, their forecasting process still often collapses into a form of \emph{narrative aggregation}: as illustrated in \cref{fig:baseline_case}, they list various pieces of information, weigh them qualitatively in prose, and then attach a probability estimate without an explicit computational pathway from evidence to forecast. As a result, it remains unclear how the final forecast probability is actually derived. This creates two problems. First, qualitative balancing can yield suboptimally calibrated probability estimates, especially when a question admits a natural quantitative anchor. Second, because the intermediate reasoning objects are not made explicit, forecasting errors and update decisions become difficult to inspect systematically after event resolution.

To address these issues, we propose \textsc{AuditForecast}, an agentic scaffold for forecasting built around a structured probabilistic decomposition. Instead of directly assigning a final probability through narrative synthesis alone, \textsc{AuditForecast} first attempts to anchor the forecast with a suitable quantitative baseline model. It then uses model-guided data retrieval to gather the inputs required by that model and derive a base probability, thereby preventing the agent from drifting into arbitrary and unbounded search. On top of this baseline, \textsc{AuditForecast} identifies situational factor updates outside the model's scope and combines them mechanically in odds space. This turns forecasting from an implicit prose-based judgment into a structured probabilistic process with explicit intermediate objects. As a result, \textsc{AuditForecast} not only improves forecasting quality, but also produces an auditable forecasting report that supports systematic post hoc analysis. We summarize our contributions as follows.
\begin{itemize}[leftmargin=1.3em,itemsep=2.5pt,topsep=2.5pt,parsep=0pt,partopsep=0pt]
\item We identify a structural limitation of existing agentic forecasting: they often rely on narrative aggregation, lacking an interpretable computational path from retrieved evidence to the final forecast. We address this with \textsc{AuditForecast}, a forecasting-specific scaffold that decomposes each forecast into a model-derived base probability and situational factor updates.
\item Across multiple live forecasting benchmarks and frontier foundation models, including GPT-5.4, GPT-5.4-mini, and Gemini-3-Flash, \textsc{AuditForecast} consistently improves forecasting accuracy over strong agentic baselines, surpasses market-implied references, and outperforms substantially more expensive deep-research agents while remaining Pareto-dominant in the cost--accuracy tradeoff. 
\item By producing an auditable forecasting report, \textsc{AuditForecast} supports post hoc analysis of forecast construction. This enables systematic auditing of both base-probability formation and factor-level updating, yielding new insight into LLM forecasting behavior.
\end{itemize}
\section{Related Work}

\textbf{LLM forecasting.}
Early work on LLM forecasting primarily studies standalone LLM systems. \citet{tournament} show that GPT-4 underperforms the human crowd in a real-world forecasting tournament, while \citet{wisdom} find that aggregating forecasts across multiple LLMs substantially improves performance. Related work also explores retrieval-augmented pipelines and prompt design, showing that better information access can improve forecasting accuracy \citep{human-level,news}, while prompting alone appears to yield only limited gains \citep{prompt}. Training-based approaches push further through stronger supervision, including self-supervised fine-tuning \citep{human-level}, self-play with DPO \citep{self-play,dpo}, outcome-based reinforcement learning \citep{outcome}, and large-scale reinforcement learning on synthetic forecasting questions derived from news \citep{scaling}. Together, these works establish a promising foundation for LLM-based forecasting.

\begin{figure*}[t]
\begin{center}
\centerline{\includegraphics[width=1.0\linewidth]{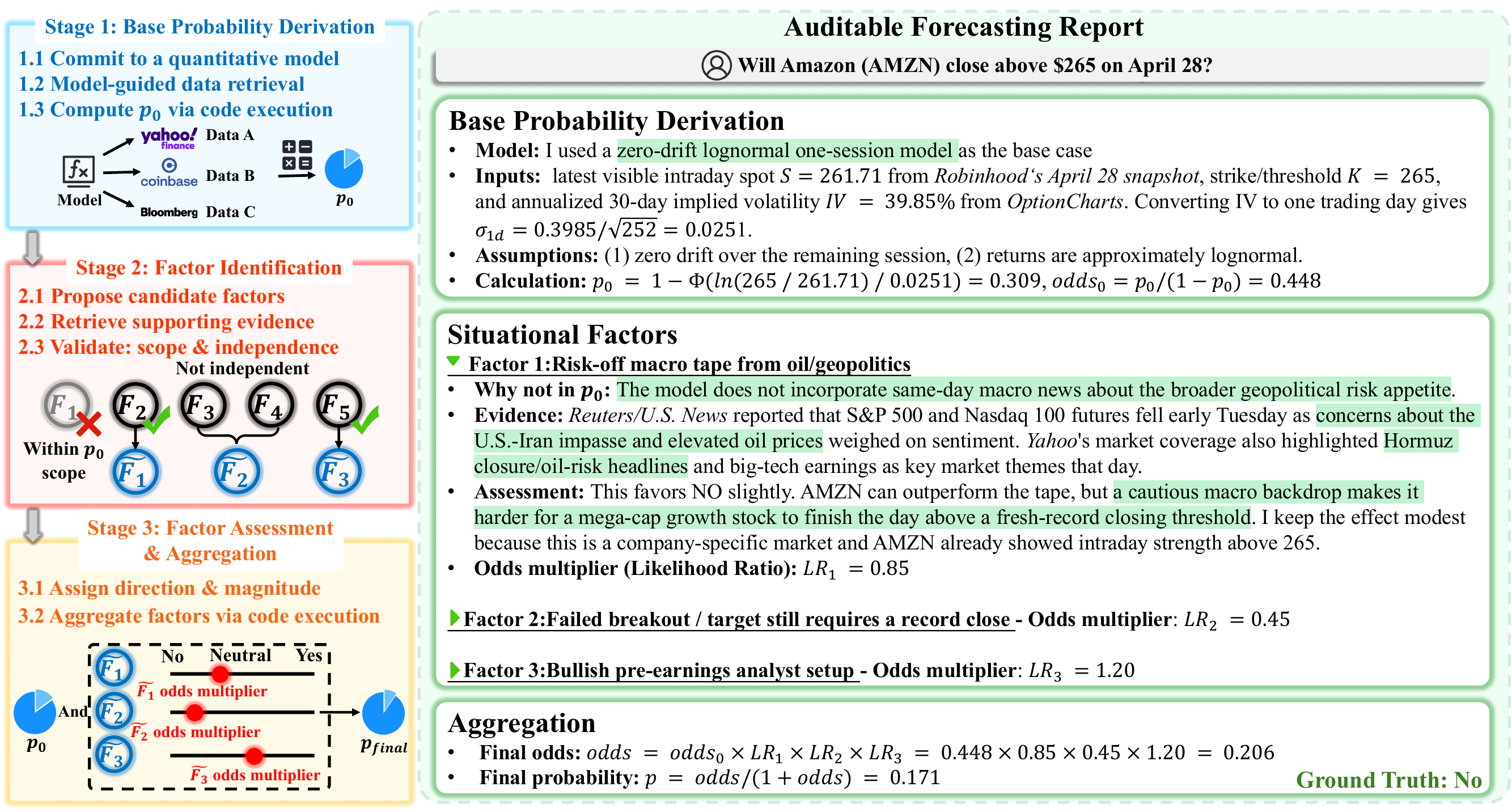}}
\vskip -0.05in
\caption{\textbf{Left:} the agent workflow under \method{}. \textbf{Right:} the corresponding auditable forecasting report. Highlighted text emphasizes key information and steps in the forecasting process.}
\label{fig:overview_case}
\end{center}
\vskip -0.35in
\end{figure*}
\noindent\textbf{Live forecasting benchmarks for LLM agents.} Recent work has introduced live forecasting benchmarks for LLM agents. ForecastBench \citep{forecastbench} and ProphetArena \citep{prophet} are built on public prediction platforms and evaluate probabilistic forecasting across diverse unresolved real-world events \citet{futurex} extend this direction to LLM agents, introducing a large-scale live benchmark for future prediction that evaluates forecasting together with search, planning, and tool use in dynamic environments. These benchmarks provide an important evaluation substrate for forecasting agents. By contrast, our work focuses on the agentic forecasting scaffold itself, aiming to make the mapping from evidence to probability explicit, auditable, and better calibrated.
\section{Methodology}\label{sec:method}

\subsection{Overview} \label{sec:method:overview}
Existing agentic forecasting is typically built on multi-turn interaction frameworks such as Think\&Search and deep research \citep{futurex}. Their common structure consists of iterative Reason--Act--Observe cycles \citep{react}, through which the agent incrementally acquires information from the internet and refines its forecast. Deep-research agents \citep{deepresearch} extend this pattern with adaptive planning and specialized search subagents, enabling more comprehensive information gathering and reasoning. Although these frameworks improve forecasting performance over base LLMs, we find that agentic forecasting still often falls into a form of \emph{narrative aggregation}: agents gather substantial information, synthesize it in prose, and then assign a final probability without an explicit mapping from evidence to probabilistic contribution, leaving the resulting forecast difficult to interpret and audit.

A large class of real-world forecasting questions, however, admits a natural quantitative anchor. For example, a stock threshold event can be anchored by a return-based model~\citep{return-model}. Such models capture the expected trajectory of the world under ordinary conditions and provide a natural probabilistic baseline. Yet no model captures every current, case-specific development relevant to the forecast. A sudden geopolitical event, for instance, may shift market sentiment in ways that fall outside the scope of the baseline model. Forecasting therefore requires combining two complementary sources of information: a model-derived baseline and evidence-grounded updates outside its scope. Motivated by this decomposition, we propose \method{}, an agentic scaffold that represents each forecast through two explicit components: a base probability \(p_0\) derived from the expected trajectory, and a set of situational factor updates outside the scope of \(p_0\). These components are combined mechanically in odds space. By enforcing this decomposition as a schema-level output constraint, \method{} produces an auditable forecasting report that makes forecast construction explicit and interpretable. In \cref{sec:method:bayes}, we formalize this decomposition by interpreting \(p_0\) as the base probability under the expected trajectory and each factor as an odds-space update. In \cref{sec:method:scaffold}, we then describe the core components of the \method{} scaffold. Detailed scaffold implementation is provided in Appendix~\ref{app:implement}.

\subsection{Structured Forecast Decomposition}
\label{sec:method:bayes}

\method{} is motivated by a Bayesian odds-space view of forecasting. We focus on binary event forecasting in this work, consistent with the formulation commonly used in prediction-market settings~\citep{prophet}. Let \(Y \in \{\mathrm{YES},\mathrm{NO}\}\) denote the event outcome, and let \(E=(E_1,\dots,E_n)\) denote the evidence associated with a set of validated factors. By Bayes' rule,
\begin{align}
\scalebox{0.8}{$
P(Y=\mathrm{YES}\mid E)
$}
&=
\scalebox{0.8}{$
\frac{P(E\mid Y=\mathrm{YES})P(Y=\mathrm{YES})}{P(E)}
$},
\label{eq:bayes_yes}
\\
\scalebox{0.8}{$
P(Y=\mathrm{NO}\mid E)
$}
&=
\scalebox{0.8}{$
\frac{P(E\mid Y=\mathrm{NO})P(Y=\mathrm{NO})}{P(E)}
$}.
\label{eq:bayes_no}
\end{align}
Taking the ratio of \cref{eq:bayes_yes,eq:bayes_no} yields
\begin{equation}
\scalebox{0.7}{$
\underbrace{\frac{P(Y=\mathrm{YES}\mid E)}{P(Y=\mathrm{NO}\mid E)}}_{\text{posterior odds}}
=
\underbrace{\frac{P(Y=\mathrm{YES})}{P(Y=\mathrm{NO})}}_{\text{prior odds}}
\cdot
\underbrace{\frac{P(E\mid Y=\mathrm{YES})}{P(E\mid Y=\mathrm{NO})}}_{\text{likelihood ratio}}.$}
\label{eq:bayes_odds}
\end{equation}
Thus, posterior odds equal prior odds multiplied by a likelihood-ratio term. If the evidence variables \(E_1,\dots,E_n\) are conditionally independent given the event outcome, the joint likelihood ratio factorizes across factors (see Appendix~\ref{app:bayes_derivation} for details):
\begin{equation}
\scalebox{1.0}{$
\frac{P(E\mid Y=\mathrm{YES})}{P(E\mid Y=\mathrm{NO})}
=
\prod_{i=1}^{n}
\frac{P(E_i\mid Y=\mathrm{YES})}{P(E_i\mid Y=\mathrm{NO})}.$}
\label{eq:factorized_lr}
\end{equation}
This motivates decomposing a forecast into a prior belief over the event and a sequence of situational factor updates. In our scaffold, the agent first uses quantitative modeling to derive a base probability \(p_0\), with corresponding odds \(o_0 = \frac{p_0}{1-p_0}\), and then identifies each situational factor and approximates its effect on the prior odds, representing that effect as an odds multiplier, denoted by \(\mathrm{LR}_i\), in likelihood-ratio form:
\begin{equation}
\scalebox{1.0}{$
\mathrm{LR}_i = \frac{P(E_i \mid Y=\mathrm{YES})}{P(E_i \mid Y=\mathrm{NO})},$}
\label{eq:lr}
\end{equation}
Under the conditional-independence assumption in \cref{eq:factorized_lr}, the final posterior odds and forecast probability follow from \cref{eq:bayes_odds}:
\begin{equation}
\scalebox{1.0}{$
o = o_0 \prod_{i=1}^{n} \mathrm{LR}_i,\quad p = \frac{o}{1+o}.$}
\label{eq:odds_aggregation}
\end{equation}
This decomposition underlies \method{}'s auditability: \(p_0\) is recorded explicitly, each factor contributes a single quantitative update term, and the final forecast \(p\) is obtained through a mechanical aggregation step.

\begin{table*}[t]
  \begin{center}\resizebox{0.9\linewidth}{!}{
    \begin{tabular}{l c c c c c c c c}
      \toprule
      \multirow{3}{*}{\raisebox{0.9ex}{\textbf{Method}}}
      & \multicolumn{4}{c}{\textbf{Polymarket}}
      & \multicolumn{4}{c}{\textbf{Kalshi}} \\
      \cmidrule(lr){2-5} \cmidrule(lr){6-9}
      & \textbf{Stocks} & \textbf{Crypto} & \textbf{Sports} & \textbf{Overall}
      & \textbf{Finance} & \textbf{Entertain.} & \textbf{Sports} & \textbf{Overall} \\

      \midrule
      \rowcolor{gray!20}\multicolumn{9}{c}{\textit{\textbf{GPT-5.4}}} \\
      Think\&Search    & 0.1063 & 0.1599 & 0.2538 & 0.1342 & 0.0588 & 0.1276 & 0.2258 & 0.1028 \\
      Market-Implied   & 0.1265 & 0.1582 & 0.2540 & 0.1494 & 0.0759 & 0.1073 & 0.2263 & 0.1060 \\
      \textbf{AuditForecast}
                       & \textbf{0.0950} & \textbf{0.1257} & \textbf{0.2517} & \textbf{0.1225}
                       & \textbf{0.0524} & \textbf{0.0943} & \textbf{0.2119} & \textbf{0.0870} \\
      \midrule
      \rowcolor{gray!20}\multicolumn{9}{c}{\textit{\textbf{GPT-5.4-mini}}} \\
      Think\&Search    & 0.1116 & 0.1531 & 0.2600 & 0.1387 & 0.0704 & 0.1862 & 0.2296 & 0.1280 \\
      Market-Implied   & 0.1265 & 0.1582 & 0.2540 & 0.1494 & 0.0759 & \textbf{0.1073} & 0.2263 & \textbf{0.1060} \\
      \textbf{AuditForecast}
                       & \textbf{0.1021} & \textbf{0.1240} & \textbf{0.2465} & \textbf{0.1270}
                       & \textbf{0.0567} & 0.1555 & \textbf{0.2041} & 0.1074 \\
      \midrule
      \rowcolor{gray!20}\multicolumn{9}{c}{\textit{\textbf{Gemini-3-flash}}} \\
      Think\&Search    & 0.1197 & 0.2252 & 0.2477 & 0.1603 & 0.0972 & 0.2458 & 0.2549 & 0.1699 \\
      Market-Implied   & 0.1251 & 0.2071 & 0.2452 & 0.1617 & 0.0697 & \textbf{0.1143} & 0.2588 & \textbf{0.1166} \\
      \textbf{AuditForecast}
                       & \textbf{0.0982} & \textbf{0.2021} & \textbf{0.2398} & \textbf{0.1420}
                       & \textbf{0.0651} & 0.1695 & \textbf{0.2441} & 0.1284 \\
      \bottomrule
    \end{tabular}}
  \end{center}
\vskip -0.1in
  \caption{Brier score ($\downarrow$) on Polymarket and Kalshi. The Market-Implied reference for Gemini-3-flash differs from that of the GPT backbones because it is evaluated on a different live time range.}
\label{tab:main_poly_kalshi}
\vskip -0.2in
\end{table*}

\subsection{The \method{} Scaffold}
\label{sec:method:scaffold}

We now operationalize the forecast decomposition in \cref{sec:method:bayes} as an agentic scaffold. Given a forecasting question, \method{} organizes the agent's behavior into three explicit stages: deriving a base probability \(p_0\), identifying model-external factors, and assessing and aggregating factor updates. While the underlying agent may iterate within a stage through search, reasoning, and computation, the scaffold constrains the final output to a structured forecasting report so that the path from evidence to probability remains explicit and auditable. \Cref{fig:overview_case} (left) illustrates the corresponding agent workflow under the scaffold.

\noindent\textbf{Stage 1: Base Probability Derivation.} The first stage begins by establishing the forecasting context through lightweight search, including the current state of the question, the relevant entities, and the resolution horizon. The scaffold then requires the agent to assess whether the question admits a suitable quantitative model for baseline estimation. If so, the agent commits to that model and explicitly records the model choice, the required inputs, and the key assumptions. This commitment induces model-guided data retrieval: instead of allowing the baseline estimate to be shaped by arbitrary search trajectories, subsequent search is directed toward the quantities required to instantiate the chosen model. In practice, this steers the agent toward primary sources for grounded inputs, such as price and options data from Yahoo Finance or official earnings information from company investor-relations pages, reducing reliance on third-party speculation, commentary, and other search artifacts that can distort the baseline estimate. The retrieved inputs are then passed to code execution to enable accurate computation of \(p_0\). The output of this stage is thus not only a base probability \(p_0\), but also a structured record of how that baseline was constructed.

\noindent\textbf{Stage 2: Factor Identification.} The second stage identifies model-external factors that should update the base probability. In existing agentic forecasting, agents often respond to retrieved evidence by listing loosely organized pros and cons, without explicitly distinguishing information already encoded in the baseline model from information that should enter only as a later update. As a result, the update process becomes vulnerable to double counting and overlapping factors. \method{} addresses this problem by making factor identification conditional on the constructed baseline \(p_0\). Rather than collecting additional evidence indiscriminately, the agent asks what relevant information is still not encoded by the baseline model and proposes candidate factors, such as broader macroeconomic, political, or market-wide conditions. It then extends search along these evidence channels to instantiate each candidate factor with concrete supporting evidence. Each instantiated factor is subsequently validated along two dimensions. First, the agent performs a scope check, verifying that the factor is genuinely outside the scope of \(p_0\) rather than a restatement of information already encoded in the baseline model. Second, the agent performs an independence check: because the odds-space aggregation in \cref{sec:method:bayes} assumes conditional independence across factors, candidates that share the same underlying causal mechanism are further revised, merged, or discarded. The output of this stage is a structured set of validated factors that are evidence-grounded, out of scope of \(p_0\), and sufficiently distinct for subsequent aggregation.

\noindent\textbf{Stage 3: Factor Assessment and Aggregation.} The third stage assigns each validated factor an explicit quantitative role in the final forecast. For each factor \(i\), the agent records an assessment that reasons from the supporting evidence to the factor's directional impact, and then assigns an odds multiplier \(\mathrm{LR}_i\). Its interpretation is
\[
\begin{cases}
\mathrm{LR}_i > 1, & \text{favors YES},\\
\mathrm{LR}_i < 1, & \text{favors NO},\\
\mathrm{LR}_i = 1, & \text{is neutral}.
\end{cases}
\]
In this way, each factor is represented by a single scalar update term that approximates the likelihood-ratio update in \cref{eq:lr} for subsequent aggregation. Once \(p_0\) and \(\{\mathrm{LR}_i\}_{i=1}^n\) are available, the final probability is computed according to \cref{eq:odds_aggregation}.

\noindent\textbf{Auditable Forecasting Report.} The right side of \cref{fig:overview_case} shows the forecasting report produced by \method{}. This report records the base-probability derivation, the situational factors with their supporting evidence and scope justifications, and the final odds-space aggregation that yields the forecast probability. By making these intermediate objects explicit, the scaffold supports direct inspection of how the forecast was constructed and how each retained factor contributed to the final prediction. It also enables clearer post hoc analysis once the event is resolved, providing a more transparent view of LLM-agent forecasting behavior. More case studies are provided in Appendix \ref{app:case_study}.

\section{Experiments}
\subsection{Experimental Settings}
\textbf{Agent Backbones and Implementation.}
We primarily consider three frontier LLM backbones: GPT-5.4, GPT-5.4-mini, and Gemini-3-flash. We implement the agentic scaffold using the OpenAI Agents SDK~\citep{openai-agents-sdk} and Google's Agent Development Kit~\citep{google-adk}. Detailed implementation details  are provided in Appendix~\ref{app:implement}.

\noindent\textbf{Benchmarks and Evaluation.}
We evaluate on three live forecasting benchmarks: two prediction markets, Polymarket~\citep{polymarket} and Kalshi~\citep{kalshi} (accessed through ProphetArena \citep{prophet}), and one multi-source benchmark, FutureX~\citep{futurex}. Our evaluation covers questions resolving in April--May 2026, yielding 831 questions for GPT-5.4 and GPT-5.4-mini, and 709 questions for Gemini-3-flash. The difference reflects separate evaluation runs conducted at different times, which resulted in different question pools. Detailed dataset construction is provided in Appendix~\ref{app:dataset}. As the metric, following standard practice in LLM forecasting~\citep{human-level,prophet}, we report the Brier score $\mathrm{Brier} = \frac{1}{N} \sum_{i=1}^{N} (p_i - y_i)^2$, where $p_i \in [0, 1]$ is the forecast for question $i$ and $y_i \in \{0, 1\}$ is the realized outcome. Smaller scores indicate better forecasts.

\noindent\textbf{Baselines.} 
We compare against four types of baselines. \textbf{Think\&Search} is the default agentic scaffold used by both the ProphetArena Leaderboard~\citep{prophetarena-leaderboard} and FutureX as a frontier-LLM configuration for forecasting tasks. We also compare against \textbf{deep-research agents}, which represent an advanced class of systems for open-ended information gathering and reasoning. In particular, we consider the closed-source o3-deep-research and o4-mini-deep-research \citep{deepresearch}, as well as the open-source smolagents-deep-research \citep{smolagents-open-dr}; detailed comparisons are presented in \cref{sec:results:deepresearch}. We also include \textbf{Market-Implied}, defined as the probability derived from prediction-market prices at each question's \emph{run-time} (i.e., when the agent produced its forecast). It captures the aggregate forecast of market participants and serves as a reference point. Finally, we compare against two \textbf{Time-Series Foundation Models (TSFMs)}, Chronos-2~\citep{chronos} and TimesFM~\citep{TimesFM}, on the value-threshold forecasting subset from Polymarket.
\begin{table}[t]
    \begin{center}\resizebox{0.9\linewidth}{!}{
    \begin{tabular}{l c c c}
      \toprule
      \textbf{Method}
      & \textbf{Level 1}
      & \textbf{Level 2}
      & \textbf{Overall} \\
      \midrule
      \rowcolor{gray!20}\multicolumn{4}{c}{\textit{\textbf{GPT-5.4}}} \\
      Think\&Search           & 0.2879          & 0.1091          & 0.1274 \\
      \textbf{AuditForecast}  & \textbf{0.0932} & \textbf{0.0921} & \textbf{0.0922} \\
      \midrule
      \rowcolor{gray!20}\multicolumn{4}{c}{\textit{\textbf{GPT-5.4-mini}}} \\
      Think\&Search           & 0.0970          & 0.1251          & 0.1223 \\
      \textbf{AuditForecast}  & \textbf{0.0965} & \textbf{0.1239} & \textbf{0.1211} \\
      \midrule
      \rowcolor{gray!20}\multicolumn{4}{c}{\textit{\textbf{Gemini-3-flash}}} \\
      Think\&Search           & 0.0728          & 0.1837          & 0.1671 \\
      \textbf{AuditForecast}  & \textbf{0.0675} & \textbf{0.1438} & \textbf{0.1324} \\
      \bottomrule
    \end{tabular}}
  \end{center}
\vskip -0.1in
  \caption{Brier score ($\downarrow$) on FutureX. }
  \label{tab:main_futurex}
\end{table}

\subsection{Main Results} \label{sec:main_results}
\noindent\textbf{AuditForecast consistently outperforms Think \&Search with the same backbone and enables frontier LLMs to surpass market-implied references.} \Cref{tab:main_poly_kalshi} and \cref{tab:main_futurex} summarize the live evaluation results on public prediction platforms (Polymarket and Kalshi) and on FutureX, respectively. Across all backbones and benchmarks, \textsc{AuditForecast} yields consistent gains over the Think\&Search baseline. On Polymarket, every backbone under \textsc{AuditForecast} surpasses the market-implied reference, while on Kalshi GPT-5.4 also exceeds the market. The gains also transfer to smaller models: for example, GPT-5.4-mini under \textsc{AuditForecast} outperforms GPT-5.4 under the baseline scaffold on Polymarket. These results suggest that the scaffold unlocks additional forecasting capability from the underlying base models.

\begin{figure}[t]
\begin{center}
\centerline{\includegraphics[width=1.0\linewidth]{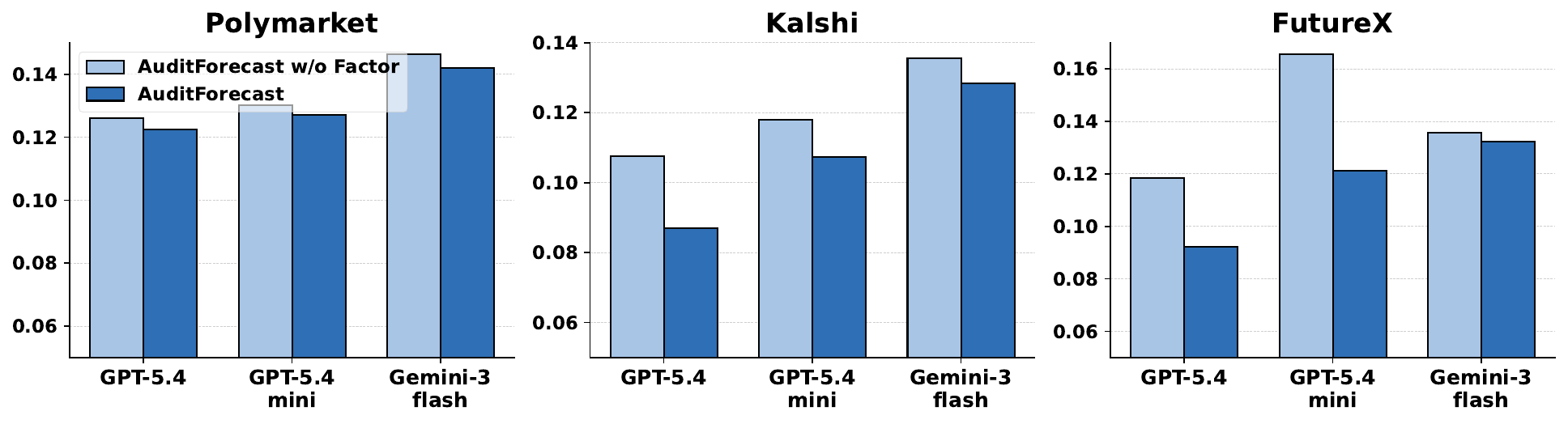}}
\vskip -0.07in
\caption{Brier score ($\downarrow$) of \textsc{AuditForecast}
with and without factor updates. Factor updates consistently improve forecasting performance.}
\label{fig:factor_update}
\end{center}
\vskip -0.2in
\end{figure}

\noindent\textbf{Factor updates further calibrate the base probability and improve forecasting accuracy.} \Cref{fig:factor_update} reports the Brier scores of \textsc{AuditForecast} with and without factor updates across three backbones on Polymarket, Kalshi, and FutureX. In all settings, including factor updates improves performance, demonstrating the value of the structured forecast decomposition in refining the base probability with evidence-grounded adjustments.

\subsection{Effect of Varying Reasoning Effort} \label{sec:effort}
\begin{figure}[t]
\begin{center}
\centerline{\includegraphics[width=1.0\linewidth]{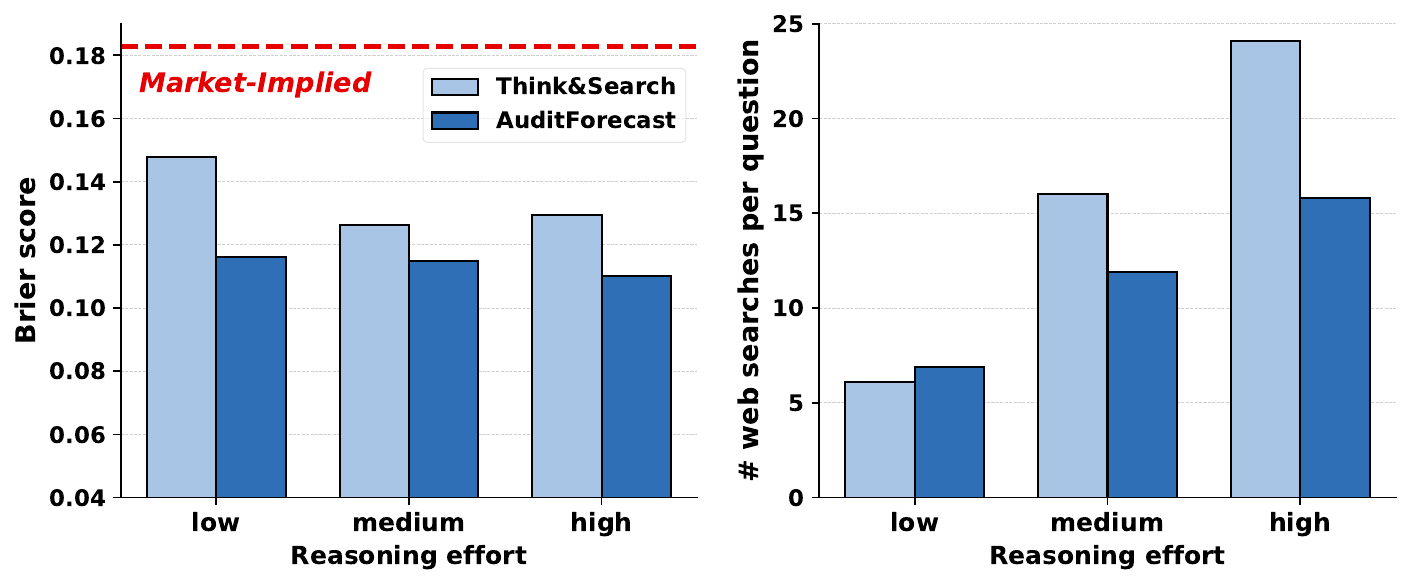}}
\vskip -0.1in
\caption{Effect of reasoning effort on forecasting accuracy and search usage. Results are shown for GPT-5.4 on Polymarket questions.}
\label{fig:effort}
\end{center}
\vskip -0.2in
\end{figure}

Frontier LLMs allow the amount of reasoning effort at inference time to be adjusted. In \cref{sec:main_results}, GPT-5.4 and GPT-5.4-mini use medium reasoning effort by default. Here, we further examine how varying reasoning effort affects forecasting performance through a focused evaluation on 90 Polymarket questions resolving in May 2026, using GPT-5.4 as the backbone. As shown in \cref{fig:effort} (left), \textsc{AuditForecast} outperforms Think\&Search at low, medium, and high reasoning effort, and yields substantially lower Brier scores than the market-implied reference across all three settings. The gap is especially pronounced at low reasoning effort, where Think\&Search degrades markedly while \textsc{AuditForecast} remains comparatively stable, suggesting that the structured decomposition in \textsc{AuditForecast} reduces the dependence of forecasting quality on larger reasoning budgets. \Cref{fig:effort} (right) reports the average number of web searches per question. At medium and high reasoning effort, \textsc{AuditForecast} uses substantially fewer searches than Think\&Search while still achieving better forecasting accuracy. This pattern is consistent with the scaffold's model-guided search behavior: rather than allowing search to expand arbitrarily as reasoning effort increases, \textsc{AuditForecast} directs search toward model inputs and factor-relevant evidence. Overall, these results show that \textsc{AuditForecast} is not only more accurate, but also more robust to reasoning budget and more targeted in its use of search.

\subsection{Comparison with Deep Research Agents}
\label{sec:results:deepresearch}
\begin{figure}[t]
\begin{center}
\centerline{\includegraphics[width=0.995\linewidth]{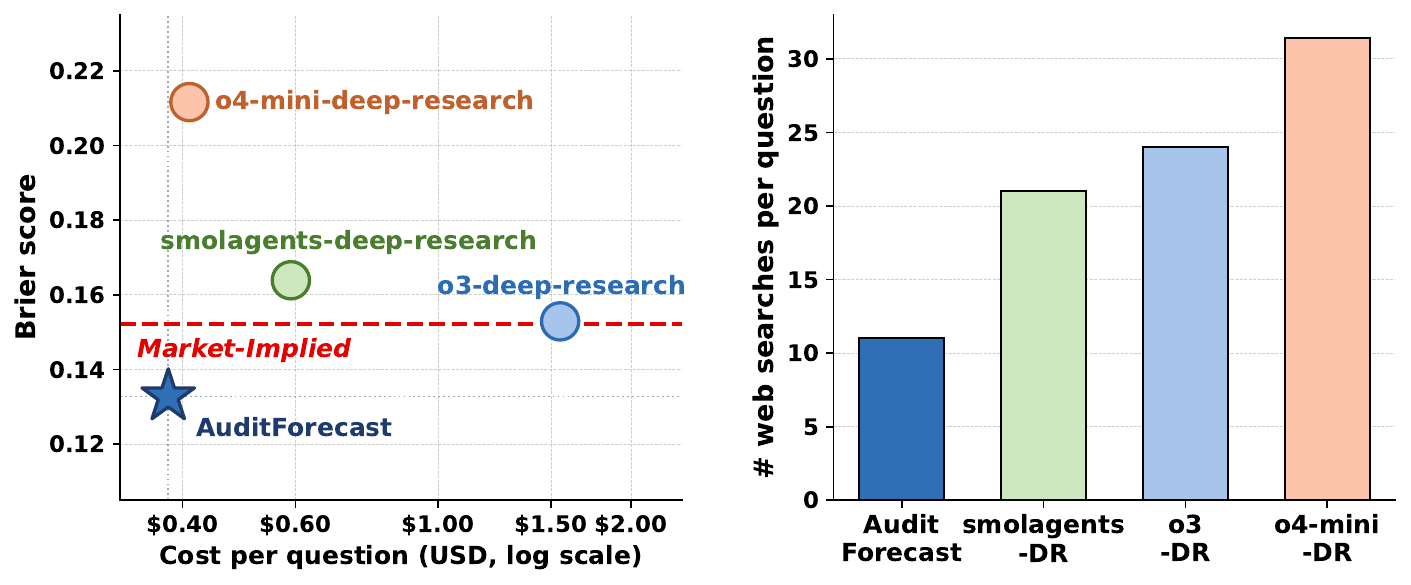}}
\vskip -0.098in
\caption{Comparison with deep research agents. \textbf{Left:} Brier score ($\downarrow$) versus cost per question. \textbf{Right:} average number of web searches per question.}
\label{fig:dr_compare}
\end{center}
\vskip -0.2in
\end{figure}

Deep research agents are typically designed for complex research topics: they can plan their own research strategy, iteratively browse dozens of pages, and ultimately compile comprehensive reports \citep{deepresearch-bench}. In this section, we further compare \method{} against such systems. Specifically, we consider OpenAI's flagship o3-deep-research and o4-mini-deep-research \citep{deepresearch}, as well as the open-source smolagents-deep-research~\citep{smolagents-open-dr} from Huggingface. For smolagents-deep-research and \method{}, we use GPT-5.4 as the foundation model. We evaluate all methods on the same Polymarket evaluation set, consisting of 142 forecasting questions resolving in May 2026.  As shown in \cref{fig:dr_compare} (left), \method{} achieves the best overall forecasting performance while also operating at the lowest cost per question, making it the unique Pareto-dominant method in the cost--accuracy plane. It outperforms all three deep-research agents in Brier score and also improves over the Market-Implied reference. \Cref{fig:dr_compare} (right) further shows that \method{} uses substantially fewer web searches per question than the deep-research agents. Thus, the performance gain does not come from broader search, but from more targeted and forecasting-specific use of search, consistent with the trend observed in \cref{sec:effort}. We observe that even deep research agents remain fundamentally narrative in their forecasting behavior. They gather more information and produce longer syntheses, but the final probability is still typically attached to the end of a prose-based discussion rather than derived through an explicit probabilistic decomposition. We provide additional qualitative comparisons in Appendix \ref{app:case_study}.

\subsection{Comparison with Time Series Forecasting}
\label{sec:results:tsfm}

Our evaluation setting follows that of prediction markets: the agent is given only a forecasting question and its resolution rule, rather than a large volume of pre-processed time-series data. The task therefore primarily evaluates the agent's ability to autonomously search for relevant information and reason about it. Time-Series Foundation Models (TSFMs), in contrast, typically require historical time-series data as input. To enable a comparison with this class of methods, we therefore prepare the relevant one-year historical time series for each of the 369 value-threshold questions in the Polymarket subset and evaluate Chronos-2 \citep{chronos} and TimesFM \citep{TimesFM} against \method{} with GPT-5.4. For each question, we derive the event probability from the predictive quantiles produced by the TSFM. As shown in \cref{tab:tsfm_comparison}, \method{} outperforms both TSFMs. TSFMs are restricted to modeling a fixed historical time series, whereas \method{} is more flexible in both information acquisition and model construction. Notably, the base probability produced by \method{} alone already surpasses both TSFMs. Moreover, prediction-market events are often driven by real-time developments and news, which purely time-series-based models cannot capture.

\begin{table}[H]
\vskip -0.02in
    \begin{center}\resizebox{0.8\linewidth}{!}{
    \begin{tabular}{l c}
      \toprule
      \textbf{Method}
      & \textbf{Threshold Forecast} \\
      \midrule
      Chronos-2               & 0.1237 \\
      TimesFM                 & 0.1241 \\
      \textbf{AuditForecast $p_0$}    & 0.0980 \\
      \textbf{AuditForecast}        & \textbf{0.0938} \\
      \bottomrule
    \end{tabular}}
  \end{center}
\vskip -0.1in
  \caption{Brier score ($\downarrow$) on value-threshold forecasts.}
  \label{tab:tsfm_comparison}
\vskip -0.1in
\end{table}

\subsection{Component-Level Ablation Study}
\label{sec:ablation}
In this section, we conduct component-level ablations using GPT-5.4 on 37 Polymarket forecasting events that resolved by July 10.

\noindent\textbf{Model-guided retrieval.} We compare our model-guided retrieval strategy against a retrieval-then-modeling variant, which first performs open-ended retrieval and then constructs a quantitative model based on the retrieved information. As shown in \cref{tab:retrieval_ablation}, model-guided retrieval requires substantially fewer web searches while producing a more accurate estimate of the base probability $p_0$. Without an explicit model to guide the search, retrieval tends to collect more redundant or irrelevant information. Moreover, because the subsequent model is constrained by the information obtained during this unguided retrieval stage, it is more likely to yield a suboptimal quantitative baseline. These results support our design choice of constructing the model first and using its information requirements to guide targeted retrieval.

\begin{table}[t]
    \begin{center}\resizebox{1.0\linewidth}{!}{
    \begin{tabular}{l c c}
      \toprule
      \textbf{Method}
      & \textbf{Avg. Web Searches}
      & \textbf{$p_0$ Brier ($\downarrow$)} \\
      \midrule
      \textbf{Model-guided retrieval}
      & \textbf{11.7}
      & \textbf{0.0802} \\
      Retrieval-then-modeling
      & 23.9
      & 0.0882 \\
      \bottomrule
    \end{tabular}}
    \end{center}
    \vskip -0.1in
    \caption{Ablation of model-guided retrieval.}
    \label{tab:retrieval_ablation}
\end{table}

\noindent\textbf{Scope / independence checking and odds aggregation.} We further ablate scope checking and independence checking individually. To evaluate the aggregation mechanism, we compare odds-space aggregation against two alternatives: (1) \textit{narrative aggregation}, where the agent directly integrates all factor evidence into a final probability without an explicit aggregation rule, and (2) \textit{probability-space aggregation}, which applies signed additive adjustments directly in probability space, $p = p_0 + \sum_i \delta_i$. As shown in \cref{tab:component_ablation}, the full \method{} pipeline achieves the best Brier score, and removing or replacing any individual component degrades forecasting performance. In particular, odds-space aggregation has several advantages over direct probability-space addition. An additive adjustment of a fixed magnitude changes the forecast equally regardless of the base probability, whereas an odds-space update naturally induces a probability shift that depends on $p_0$. Probability-space addition can also produce values outside $[0,1]$, while multiplying likelihood ratios preserves a valid probability after conversion back from odds space. These results support the use of scope checking, independence checking, and odds-space aggregation as complementary components of the forecasting scaffold.

\begin{table}[t]
    \begin{center}\resizebox{0.82\linewidth}{!}{
    \begin{tabular}{l c}
      \toprule
      \textbf{Method}
      & \textbf{Brier ($\downarrow$)} \\
      \midrule
      \textbf{\method{} }
      & \textbf{0.0764} \\
      w/o scope checking
      & 0.0817 \\
      w/o independence checking
      & 0.0824 \\
      Narrative aggregation
      & 0.0815 \\
      Probability-space aggregation
      & 0.0807 \\
      \bottomrule
    \end{tabular}}
    \end{center}
    \vskip -0.1in
    \caption{Ablation of scope checking, independence checking, and aggregation.}
    \label{tab:component_ablation}
\end{table}

\section{Auditing Forecast Construction}
In this section, we use the structured decomposition of \textsc{AuditForecast} to audit intermediate reasoning components and expose forecasting-specific behaviors hidden by unstructured scaffolds. We first examine base-probability derivation in \cref{sec:baseprob_audit}, and then analyze factor updates in \cref{sec:factor_audit}.

\subsection{Auditing Base Probability Derivation} \label{sec:baseprob_audit}
For base-probability derivation, we focus on three questions: (1) when appropriate, do agents estimate \(p_0\) through quantitative modeling? (2) for questions of the same type, do agents converge on the same modeling method? and (3) even when the modeling method is shared, do agents retrieve the same data source to instantiate it? To analyze and illustrate these questions, we focus primarily on Polymarket stock price range questions under the GPT-5.4 backbone.

\noindent\textbf{Base Probability Hierarchy Extraction.} To analyze how agents construct \(p_0\), we induce a taxonomy of \(p_0\) computations using an LLM-as-judge. Our scaffold logs each \(p_0\) in a structured form recording the modeling method and the associated data sources. We then present these records to the judge in incremental batches, allowing it to summarize and refine the taxonomy over time. The resulting hierarchy has two levels: Level~1 captures the modeling method, and Level~2 captures the data source supplying the key input. A second LLM judge then independently assigns each record to a leaf node under the finalized hierarchy. More details are provided in Appendix \ref{app:audit_p0}.

\begin{figure}[t]
\begin{center}
\centerline{\includegraphics[width=1.0\linewidth]{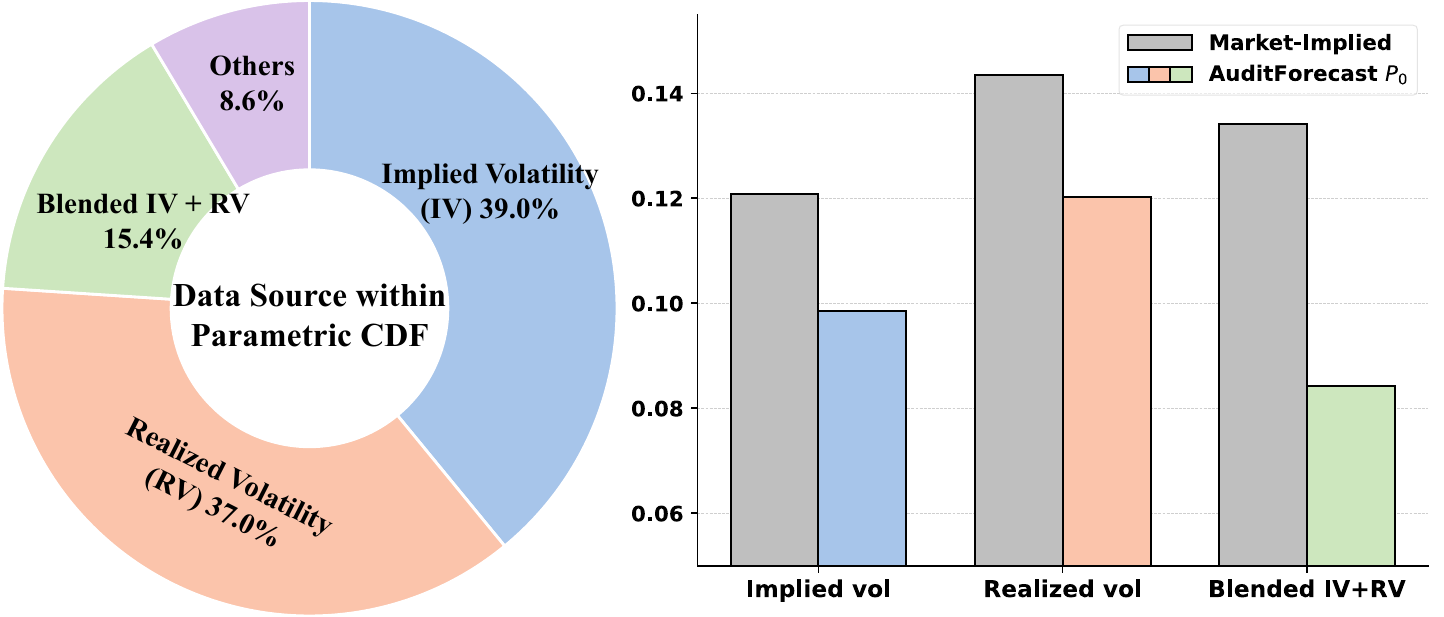}}
\vskip -0.05in
\caption{GPT-5.4 under \textsc{AuditForecast} on stock range forecasts. \textbf{Left:} Data-source distribution within the Parametric CDF method for \(p_0\). \textbf{Right:} Brier score of the dominant data-source subtypes.}
\label{fig:audit_p0}
\end{center}
\vskip -0.23in
\end{figure}

\noindent\textbf{Auditing Results.} Our statistics show that, for stock range forecasting, 91.8\% of forecasts use the \emph{Parametric CDF} method, in which the asset price at resolution is assumed to follow a parametric distribution (typically a zero-drift lognormal), and \(p_0\) is computed by evaluating the cumulative distribution function (CDF) at the question threshold. While \textsc{AuditForecast} stabilizes agents' choice of modeling method for forecasting questions of the same type, we find that agents still diverge in their choice of data source for instantiating the model. As shown in \cref{fig:audit_p0} (left), within the Parametric CDF modeling family, agents primarily split among implied volatility, realized volatility, and deliberate combinations of the two when selecting the volatility input. Furthermore, \cref{fig:audit_p0} (right) reports the Brier score of each dominant data-source subtype against the Market-Implied reference on the matched question subset. All three subtypes' \(p_0\) achieve lower Brier scores than Market-Implied, indicating that the Parametric CDF template, when instantiated with any reasonable volatility input, yields a competitive base probability.
\begin{tcolorbox}[
    enhanced,
    colback={rgb,255:red,239;green,245;blue,249},      
    colframe={rgb,255:red,47;green,111;blue,181},      
    colbacktitle={rgb,255:red,47;green,111;blue,181},  
    coltitle=white,
    title=Findings on Base-Probability Derivation,
    fonttitle=\bfseries\small,
    arc=3pt,
    boxrule=0.8pt,
    top=5pt, bottom=5pt, left=5pt, right=5pt,
    before skip=6pt,
    after skip=3pt,
    attach boxed title to top left={xshift=2mm,yshift=-2mm},
    boxed title style={
        sharp corners,
        colback={rgb,255:red,47;green,111;blue,181}, 
        colframe={rgb,255:red,47;green,111;blue,181},
        boxrule=0pt,
        top=0.3pt, bottom=0.3pt, left=3pt, right=3pt
    }
]
\textsc{AuditForecast} induces agents to use suitable quantitative models. For questions of the same type, agents tend to converge on the same modeling method while still diverging in the data source used to instantiate that model.
\end{tcolorbox}

\subsection{Auditing Factor Updates} \label{sec:factor_audit}
Once a forecasting event is resolved, we can audit factor updates by comparing the agent's factor-level judgments against the realized event trajectory. We focus on two questions: (1) whether the assigned odds multipliers have appropriate direction and magnitude, and (2) whether the agent misses important drivers of the event. As a representative subset, we again focus on stock range questions with GPT-5.4 as the backbone, where macro conditions can materially affect market dynamics. We first use an LLM judge to reconstruct the realized event trajectory between forecast run time and final resolution, and then compare it against the factor updates produced by \textsc{AuditForecast}. Additional auditing procedure details are deferred to Appendix \ref{app:audit_factor}.

\begin{figure}[t]
\begin{center}
\centerline{\includegraphics[width=1.0\linewidth]{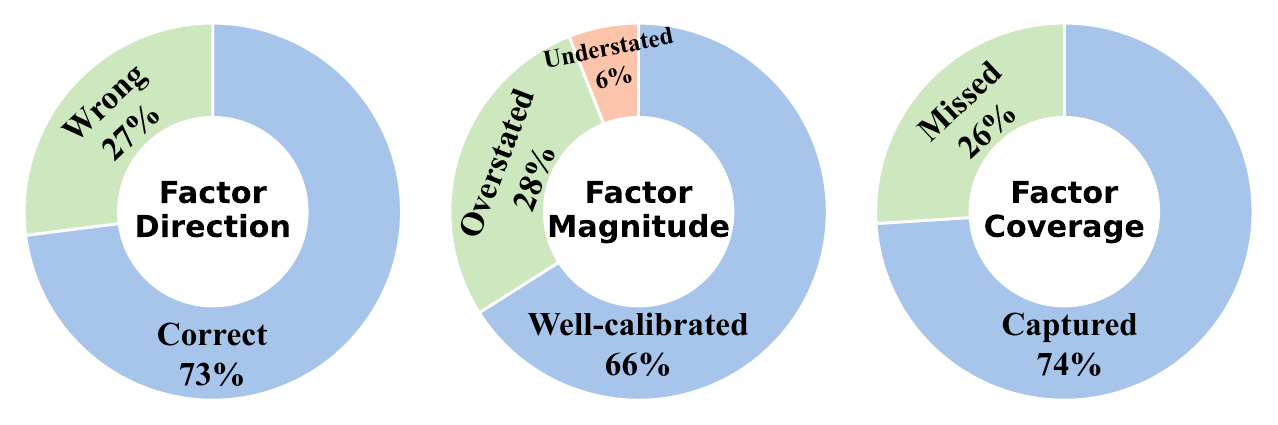}}
\vskip -0.05in
\caption{Factor auditing results for \textsc{AuditForecast} on stock range questions with GPT-5.4 as the backbone.}
\label{fig:factor_audit}
\end{center}
\vskip -0.24in
\end{figure}

\noindent\textbf{Auditing Results.} \Cref{fig:factor_audit} summarizes the factor-level auditing results. We find that \textsc{AuditForecast} with GPT-5.4 produces largely sound factor updates: 73\% of factor directions are correct, and 66\% of factor magnitudes are well calibrated. This helps explain why factor updates improve forecasting accuracy beyond the base probability derived from quantitative modeling alone. For factor magnitude, the middle plot of \cref{fig:factor_audit} shows a clear asymmetry: the GPT-5.4 agent rarely under-reacts to realized signals, while overstated factors occur substantially more often than understated ones. The post hoc judge can also identify realized factors that the agent failed to include. Here we focus only on \emph{obtainable} missing factors, i.e., factors supported by information already available at run time, since some signals emerge only after the forecast is made. As shown in \cref{fig:factor_audit} (right), the agent captures most factors (74\%), but a nontrivial fraction is still missed. We observe two specific failure modes. First, some signals are retrieved but not elevated into explicit factors. For example, across stock forecasts resolving on 2026-04-22, the agent includes an Iran-ceasefire-related macro factor in only about half of the forecasts, even though this news was a dominant driver of the broad market rally and was often mentioned in the search summary. Second, some factors require cross-source synthesis beyond a single-asset search, such as sector-level co-movement; these signals are individually retrievable but distributed across heterogeneous sources.
\begin{tcolorbox}[
    enhanced,
    colback={rgb,255:red,239;green,245;blue,249},      
    colframe={rgb,255:red,47;green,111;blue,181},      
    colbacktitle={rgb,255:red,47;green,111;blue,181},  
    coltitle=white,
    title=Findings on Factor Updates,
    fonttitle=\bfseries\small,
    arc=3pt,
    boxrule=0.8pt,
    top=5pt, bottom=5pt, left=5pt, right=5pt,
    before skip=6pt,
    after skip=3pt,
    attach boxed title to top left={xshift=2mm,yshift=-2mm},
    boxed title style={
        sharp corners,
        colback={rgb,255:red,47;green,111;blue,181},   
        colframe={rgb,255:red,47;green,111;blue,181},
        boxrule=0pt,
        top=0.3pt, bottom=0.3pt, left=3pt, right=3pt
    }
]
\textsc{AuditForecast} produces largely sound factor updates in both direction and magnitude, helping explain the gains over base-probability-only forecasting. At the same time, the GPT-5.4 agent tends to overstate some situational factors and miss some obtainable factors, especially those requiring cross-source synthesis.
\end{tcolorbox}
\section{Conclusion}
We propose \textsc{AuditForecast}, a structured scaffold for agentic forecasting that decomposes each forecast into a model-derived base probability and situational factor updates. Across multiple live benchmarks, \textsc{AuditForecast} consistently improves accuracy over strong agentic baselines and market-implied references. By producing an auditable forecasting report, it also supports systematic post hoc analysis. We hope this work provides a useful foundation for building forecasting agents that are more accurate and transparent.

\section*{Limitations}
The structured decomposition introduced by \textsc{AuditForecast} supports explicit forecast construction and produces an auditable forecasting report, which in turn enables rich post hoc analysis. In this work, such analysis already exposes several useful findings about current agentic forecasting systems. For example, as shown in Section~5, different data sources used to instantiate the same base-probability model can lead to different forecasting accuracy, and agents may systematically overlook certain classes of signals during situational factor identification. At the same time, while \textsc{AuditForecast} makes such behaviors visible, it does not yet systematically convert them into reusable improvements to the forecasting system itself. An important direction for future work is therefore to make this analysis more systematic and comprehensive: identifying recurring failure modes in model selection, data retrieval, and factor updating, and feeding these patterns back into the forecasting scaffold as reusable experience. We expect such a loop to be important for building a more self-evolving agentic forecasting system.

\section*{Ethical Considerations}
Forecasting agents may influence real-world decisions in domains such as finance, politics, and public affairs. Although \textsc{AuditForecast} improves transparency by making forecast construction explicit and auditable, it does not guarantee correctness or eliminate bias in the underlying models, retrieved information, or probabilistic judgments. We therefore view \textsc{AuditForecast} as a decision-support tool rather than a substitute for expert judgment, especially in high-stakes settings. The structured forecasting report may help users inspect how a forecast was constructed, but it should not be interpreted as a guarantee of reliability. More broadly, we believe that making forecasting outputs more explicit and auditable is an important step toward safer deployment, as it enables closer scrutiny of agent behavior and facilitates post hoc analysis when failures occur.

\section*{Acknowledgements}
We thank the anonymous reviewers for their valuable feedback and helpful suggestions.

\bibliography{custom}

\appendix
\newpage
\appendix
\section{Bayesian Derivation}
\label{app:bayes_derivation}

Let \(Y \in \{\mathrm{YES},\mathrm{NO}\}\) denote the binary event outcome, and let
\(E=(E_1,\dots,E_n)\) denote the evidence associated with a set of validated factors. Applying Bayes' rule gives
\begin{align}
\scalebox{0.9}{$
P(Y=\mathrm{YES}\mid E)
$}
&=
\scalebox{0.9}{$
\frac{P(E\mid Y=\mathrm{YES})P(Y=\mathrm{YES})}{P(E)}
$},
\label{eq:app_bayes_yes}
\\
\scalebox{0.9}{$
P(Y=\mathrm{NO}\mid E)
$}
&=
\scalebox{0.9}{$
\frac{P(E\mid Y=\mathrm{NO})P(Y=\mathrm{NO})}{P(E)}
$}.
\label{eq:app_bayes_no}
\end{align}
Taking the ratio of \cref{eq:app_bayes_yes,eq:app_bayes_no} yields
\begin{equation}
\scalebox{1.0}{$
\frac{P(Y=\mathrm{YES}\mid E)}{P(Y=\mathrm{NO}\mid E)}
=
\frac{P(E\mid Y=\mathrm{YES})}{P(E\mid Y=\mathrm{NO})}
\cdot
\frac{P(Y=\mathrm{YES})}{P(Y=\mathrm{NO})}.
$}
\label{eq:app_bayes_odds}
\end{equation}
Define the odds of the YES outcome as
\begin{equation}
\scalebox{0.9}{$
\mathrm{odds}(Y=\mathrm{YES})
=
\frac{P(Y=\mathrm{YES})}{P(Y=\mathrm{NO})}.
$}
\label{eq:app_odds_def}
\end{equation}
Then \cref{eq:app_bayes_odds} can be written as
\begin{equation}
\scalebox{0.76}{$
\mathrm{odds}(Y=\mathrm{YES}\mid E)
=
\mathrm{odds}(Y=\mathrm{YES})
\cdot
\frac{P(E\mid Y=\mathrm{YES})}{P(E\mid Y=\mathrm{NO})}.
$}
\label{eq:app_odds_update}
\end{equation}
showing that posterior odds equal prior odds multiplied by a likelihood-ratio term. If the evidence variables are conditionally independent given the event outcome, then
\begin{equation}
\scalebox{0.9}{$
P(E\mid Y=\mathrm{YES})
=
\prod_{i=1}^{n} P(E_i\mid Y=\mathrm{YES}).
$}
\label{eq:app_ci_yes}
\end{equation}
and similarly,
\begin{equation}
\scalebox{0.9}{$
P(E\mid Y=\mathrm{NO})
=
\prod_{i=1}^{n} P(E_i\mid Y=\mathrm{NO}).
$}
\label{eq:app_ci_no}
\end{equation}
Substituting \cref{eq:app_ci_yes,eq:app_ci_no} into the likelihood-ratio term in \cref{eq:app_odds_update} gives
\begin{equation}
\scalebox{0.8}{$
\begin{aligned}
\frac{P(E\mid Y=\mathrm{YES})}{P(E\mid Y=\mathrm{NO})}
&=
\frac{\prod_{i=1}^{n} P(E_i\mid Y=\mathrm{YES})}
{\prod_{i=1}^{n} P(E_i\mid Y=\mathrm{NO})}
\\
&=
\prod_{i=1}^{n}
\frac{P(E_i\mid Y=\mathrm{YES})}{P(E_i\mid Y=\mathrm{NO})}.
\end{aligned}
$}
\label{eq:app_factorized_lr}
\end{equation}
Define the factor-level likelihood-ratio term
\begin{equation}
\scalebox{0.9}{$
\mathrm{LR}_i
=
\frac{P(E_i\mid Y=\mathrm{YES})}{P(E_i\mid Y=\mathrm{NO})}.
$}
\label{eq:app_lr_def}
\end{equation}
Then the posterior odds become
\begin{equation}
\scalebox{0.82}{$
\mathrm{odds}(Y=\mathrm{YES}\mid E)
=
\mathrm{odds}(Y=\mathrm{YES})
\prod_{i=1}^{n}\mathrm{LR}_i.
$}
\label{eq:app_final_odds}
\end{equation}
In \textsc{AuditForecast}, the agent instantiates the prior odds through a quantitative-model-derived  base probability \(p_0\),
\begin{equation}
\scalebox{1.0}{$
o_0=\frac{p_0}{1-p_0}.
$}
\end{equation}
and the final odds and probability are therefore written as
\begin{equation}
\scalebox{1.0}{$
o = o_0 \prod_{i=1}^{n}\mathrm{LR}_i,
\quad
p=\frac{o}{1+o}.
$}
\label{eq:app_final_p}
\end{equation}

\section{Additional Experimental Details} \label{app:exp_setting}

\subsection{Evaluation Datasets} \label{app:dataset}
In this section, we describe the construction of the evaluation datasets in detail. To avoid contamination, all forecasting questions in our evaluation set remained unresolved at test time. We primarily draw questions from two public prediction platforms, Polymarket \citep{polymarket} and Kalshi \citep{kalshi}, as well as one forecasting benchmark, FutureX \citep{futurex}. For Polymarket, we directly collect high-attention questions from the platform website\footnote{\url{https://polymarket.com/}}. For Kalshi, we retrieve questions through the data API\footnote{\url{https://prophetarena.co/developer}} provided by ProphetArena \citep{prophet}. FutureX maintains and releases forecasting questions on a weekly basis, organized into four difficulty levels. Among them, Levels 3 and 4 contain open-ended questions. Since our work focuses on binary-event probabilities, we primarily use the Level 1 and Level 2 multiple-choice questions and convert them into independent binary forecasting problems. To prevent duplicated forecasting questions across data sources, we perform a redundancy check over the full evaluation set to remove overlaps. The overall evaluation window spans April--May 2026, and for all questions the time between forecast run time and final resolution is within one week. Because different foundation models were evaluated at different times, their resulting evaluation sets are not perfectly identical. We report the corresponding dataset statistics in \cref{tab:data_stats}.

\begin{table}[h]
\vskip -0.05in
\centering
\resizebox{0.9\linewidth}{!}{
\begin{tabular}{lcccc}
\toprule
\textbf{Model} & \textbf{Polymarket} & \textbf{Kalshi} & \textbf{FutureX} & \textbf{Total} \\
\midrule
GPT-5.4        & 562 & 220 &  49 &  831 \\
GPT-5.4-mini   & 562 & 220 &  49 &  831 \\
Gemini-3-flash & 391 & 251 &  67 &  709 \\
\bottomrule
\end{tabular}}
\vskip -0.1in
\caption{Dataset statistics.}
\label{tab:data_stats}
\vskip -0.3in
\end{table}

\subsection{Agentic Scaffold Implementation} \label{app:implement}

We implement the agentic scaffold \textsc{AuditForecast} using the OpenAI Agents SDK \citep{openai-agents-sdk} and Google's Agent Development Kit \citep{google-adk}, together with their respective hosted web-search tools, \texttt{WebSearchTool} and \texttt{google\_search}. The code is available at \url{https://github.com/CaoYuanpu/AuditForecast}. We do not impose an explicit limit on the number of search calls. Our scaffold adopts a single-agent setup: the prompt explicitly specifies the stages of the forecasting process and the objective of each stage, while an output schema is enforced to produce a structured forecasting report. Below, we provide the concrete prompts and the corresponding Pydantic implementation of the output schema. We begin with the system prompt, which provides the agent with a structured decomposition of forecasting into a base probability, situational factors, and their aggregation rules:
\begin{tcolorbox}[
    enhanced,
    colback={rgb,255:red,239;green,245;blue,249},      
    colframe={rgb,255:red,47;green,111;blue,181},      
    colbacktitle={rgb,255:red,47;green,111;blue,181},  
    coltitle=white,
    title=\textsc{AuditForecast} System Prompt,
    fonttitle=\bfseries\small,
    arc=3pt,
    boxrule=0.8pt,
    top=5pt, bottom=5pt, left=5pt, right=5pt,
    before skip=3pt,
    attach boxed title to top left={xshift=2mm,yshift=-2mm},
    boxed title style={
        sharp corners,
        colback={rgb,255:red,47;green,111;blue,181}, 
        colframe={rgb,255:red,47;green,111;blue,181},
        boxrule=0pt,
        top=0.3pt, bottom=0.3pt, left=3pt, right=3pt
    }
]
\small
\textbf{System prompt:}
You are a probabilistic forecasting analyst with web search and code
execution capabilities.\\[3pt]
Given a forecasting question, your task is to produce a calibrated
probability estimate using a structured Bayesian approach: derive a base
probability ($p_0$), identify independent factors not captured by $p_0$,
and combine them via odds multiplier updates.\\
\rule{1.0\linewidth}{0.2pt}\\[3pt]
\textbf{BASE PROBABILITY ($p_0$)}\\
$p_0$ is the baseline probability of YES under the default expected
trajectory from the current state to the resolution date, before
incorporating additional factors. To estimate $p_0$, use whatever method
is most appropriate for the question:
\begin{itemize}[leftmargin=12pt, topsep=2pt, itemsep=1pt]
\item If a quantitative model naturally applies, compute it and use it as
the primary anchor. Python execution is available via the shell for
numerical calculations.
\item If no quantitative model applies, estimate $p_0$ from the best
available qualitative assessment of the current situation.
\end{itemize}
In all cases, state the method, the data inputs, and the key
assumptions.\\
\rule{1.0\linewidth}{0.2pt}\\[3pt]
\textbf{FACTORS}\\
Factors are case-specific developments or conditions
not already captured by $p_0$ that could meaningfully shift the outcome.
Each factor must satisfy:
\begin{enumerate}[leftmargin=12pt, topsep=2pt, itemsep=1pt]
\item Outside $p_0$'s scope -- the method used to derive $p_0$
does not already account for it.
\item Supported by evidence -- concrete, current evidence exists
(found through web search).
\item Independent -- it does not share a causal mechanism with
another listed factor. If two candidate factors trace back to the same root cause, merge them into one.
\end{enumerate}
For each factor:\\
-- search for concrete evidence \\[3pt]
-- assess how the evidence shifts the odds of YES (direction and magnitude) \\[3pt]
-- express the impact as an odds multiplier ($>$1 favors YES, $<$1 favors NO, 1 = neutral).\\[3pt]
\rule{1.0\linewidth}{0.2pt}\\[3pt]
\textbf{BAYESIAN UPDATE} \\
Combine $p_0$ and all factor updates
mechanically:
\begin{align*}
o_0 &= p_0 / (1 - p_0) \\
o_{\text{final}} &= o_0 \cdot \mathrm{LR}_1 \cdots \mathrm{LR}_n \\
p_{\text{final}} &= o_{\text{final}} / (1 + o_{\text{final}})
\end{align*}
Use code execution for the arithmetic to avoid rounding errors. Show the
full calculation.
\normalsize
\end{tcolorbox}

\noindent The user prompt mainly includes the binary event forecasting question, the corresponding resolution rules specifying how the market is settled, the current time, and the expected event resolution deadline, when available.
\begin{tcolorbox}[
    enhanced,
    colback={rgb,255:red,239;green,245;blue,249},      
    colframe={rgb,255:red,47;green,111;blue,181},      
    colbacktitle={rgb,255:red,47;green,111;blue,181},  
    coltitle=white,
    title=\textsc{AuditForecast} User Prompt,
    fonttitle=\bfseries\small,
    arc=3pt,
    boxrule=0.8pt,
    top=5pt, bottom=5pt, left=5pt, right=5pt,
    attach boxed title to top left={xshift=2mm,yshift=-2mm},
    before skip=3pt,
    after skip=4pt,
    boxed title style={
        sharp corners,
        colback={rgb,255:red,47;green,111;blue,181}, 
        colframe={rgb,255:red,47;green,111;blue,181},
        boxrule=0pt,
        top=0.3pt, bottom=0.3pt, left=3pt, right=3pt
    }
]
\small
\textbf{User prompt:}\\
Forecasting question:\\
``\{\textsl{binary question}\}''\\[3pt]
Resolution criteria:\\
\{\textsl{resolution rules describing how the market is settled}\}\\[3pt]
Close date: \{\textsl{YYYY-MM-DD}\}\\
Current date: \{\textsl{YYYY-MM-DD, the run-time anchor}\}\\[6pt]
Task:
\begin{enumerate}[leftmargin=12pt, topsep=2pt, itemsep=1pt]
\item Search for relevant information.
\item Derive the base probability $p_0$ using the most appropriate
method. State the method, inputs, and assumptions. Use code execution
if a quantitative model applies.
\item Identify independent factors outside $p_0$'s scope. For each,
search for evidence and assess its impact as an odds multiplier.
\item Compute the final probability via Bayesian update:
$o_0 \cdot \prod_i \mathrm{LR}_i \to p_{\text{final}}$. Use code
execution for the arithmetic.
\end{enumerate}
Return a JSON object matching the provided schema.
\normalsize
\end{tcolorbox}

\noindent To further ensure that the agent follows the proposed forecasting workflow and ultimately produces auditable forecasting outputs, we explicitly define its output schema:
\begin{lstlisting}[style=auditpython]
from pydantic import BaseModel
class Factor(BaseModel):
    # concise factor name
    name: str
    # justification it lies outside p0's scope
    why_not_in_p0: str
    # concrete facts found via web search
    evidence: str
    # how evidence shifts odds (direction & magnitude)
    assessment: str
    # >1 favors YES, <1 favors NO, 1 = neutral
    odds_multiplier: float

class AuditForecastOutput(BaseModel):
    # overview of retrieved information
    search_summary: str
    # method, data inputs, key assumptions
    p0_reasoning: str
    # base probability of YES, in [0, 1]
    p0: float
    # base odds = p0 / (1 - p0)
    o0: float
    # independent factors outside p0's scope.
    # factors sharing a causal mechanism 
    # should be merged into a single factor.
    factors: list[Factor]
    # o0 * LR_1 * ... * LR_n = o_final trace
    bayesian_update: str
    # final probability after factor updates
    prob_yes: float
\end{lstlisting}

\section{Qualitative Case Comparison: Deep Research Agents vs.\ \textsc{AuditForecast}} \label{app:case_study}
Here we provide a detailed qualitative comparison between the forecasting reports produced by several deep-research agent systems and \textsc{AuditForecast} on the same forecasting task. Specifically, we present outputs from o4-mini-deep-research, o3-deep-research, smolagents-deep-research with GPT-5.4, and \textsc{AuditForecast} with GPT-5.4. We observe that all three deep-research agents successfully collect a wide range of relevant evidence, including distributional anchors (e.g., spot price versus threshold) and distinctions between longer-term and shorter-term signals. However, they all remain at the level of prose narrative. They identify evidence but do not quantify it, weigh evidence but do not combine it mechanically, and jump from qualitative conclusions to specific probabilities without revealing the intermediate mapping. As a result, they incur substantial search cost while still producing suboptimal forecasting accuracy. By contrast, \textsc{AuditForecast} provides an explicit path from evidence to forecast probability through a structured decomposition of base probability and factor-level updates.
\begin{tcolorbox}[
    enhanced,
    breakable,
    pad at break*=2mm,
    colback={rgb,255:red,247;green,247;blue,247},
    colframe={rgb,255:red,130;green,130;blue,130},
    colbacktitle={rgb,255:red,130;green,130;blue,130},
    coltitle=white,
    title=o4-mini-deep-research,
    title after break=o4-mini-deep-research (continued),
    fonttitle=\bfseries\small,
    arc=3pt, boxrule=0.8pt,
    top=5pt, bottom=5pt, left=5pt, right=5pt,
    attach boxed title to top left={xshift=2mm,yshift=-2mm},
    boxed title style={
        sharp corners,
        colback={rgb,255:red,130;green,130;blue,130},
        colframe={rgb,255:red,130;green,130;blue,130},
        boxrule=0pt,
        top=0.3pt, bottom=0.3pt, left=3pt, right=3pt
    }
]
\small
\textbf{Question:}\\Will Apple (AAPL) close above \$300 on May 19?\\
\textbf{Ground truth:} No\\
\textbf{Predicted probability:} 0.75\\
\textbf{Web searches:} 33\\
\textbf{Output report:}\\
Apple's stock has been on a strong upward run in recent weeks. After
its May 1 earnings beat (revenue \$111B, well above expectations) and
guidance raise, Apple announced a dividend bump and \$100B buyback,
which lifted investor sentiment. Technical indicators are also
bullish. As of mid-May, Apple was trading in a confirmed uptrend above
its key moving averages, with ``consistent higher highs since March
2026,'' and momentum (RSI) in overbought territory, reflecting
``robust bullish momentum'' and rising volume. On May 13th the stock
hit a new record intraday high of \$300, only closing ``just below
\$299'' at the day's end.\\

All this suggests strong buying interest pushing AAPL higher. In fact,
analysts like Bank of America have targets well above \$300 (around
\$325) on decently positive catalysts (new chip deals, AI strategy
coming at WWDC, expected foldable iPhone, etc.). In this context, and
absent any sudden sell-off, the odds seem favorable that the May 19
closing price will exceed \$300. (One caution is that markets can pull
back after such a run --- but there's no obvious immediate negative
catalyst.) Based on Apple's recent breakout past \$295 and the
momentum into record highs, a ``Yes'' resolution is significantly more
likely than ``No'' though not certain.\\

\textbf{FINAL\_PROBABILITY: 0.75}
\normalsize
\end{tcolorbox}

\begin{tcolorbox}[
    enhanced,
    breakable,
    pad at break*=2mm,
    colback={rgb,255:red,247;green,247;blue,247},
    colframe={rgb,255:red,130;green,130;blue,130},
    colbacktitle={rgb,255:red,130;green,130;blue,130},
    coltitle=white,
    title=o3-deep-research,
    title after break=o3-deep-research (continued),
    fonttitle=\bfseries\small,
    arc=3pt, boxrule=0.8pt,
    top=5pt, bottom=5pt, left=5pt, right=5pt,
    attach boxed title to top left={xshift=2mm,yshift=-2mm},
    boxed title style={
        sharp corners,
        colback={rgb,255:red,130;green,130;blue,130},
        colframe={rgb,255:red,130;green,130;blue,130},
        boxrule=0pt,
        top=0.3pt, bottom=0.3pt, left=3pt, right=3pt
    }
]
\small
\textbf{Question:}\\Will Apple (AAPL) close above \$300 on May 19?\\
\textbf{Ground truth:} No\\
\textbf{Predicted probability:} 0.45\\
\textbf{Web searches:} 24\\
\textbf{Output report:} \\

\textbf{Understanding the Question.}
The question asks for the probability that Apple will close above \$300
on May 19, 2026. Since the current date is May 19 and the market closes
today, we need to estimate the probability that Apple's stock will end
the day above \$300, considering recent stock price, market conditions,
news and events, analyst sentiment, and historical volatility.\\

\textbf{Current Price and Recent Performance.}
On the last trading day (May 18), Apple closed at \$297.84. Specifically,
a move of roughly \$2.16 (about +0.7\% from \$297.84) by the end of the
day would result in a ``Yes'' outcome. The May 18 close represented a
0.80\% decline from the day prior, indicating that Apple had a modest
pullback. Apple has been on an upward trajectory in 2026, reaching a
52-week high of \$303.20 in recent weeks, meaning it has already
flirted with levels just above \$300. This underscores that \$300 is a
key resistance level the stock has been testing.\\

\textbf{Pro ``Yes'':}
\begin{itemize}[leftmargin=12pt, topsep=2pt, itemsep=1pt]
  \item Small gap to \$300 (only +0.7\% needed).
  \item Upward momentum: stock has shown strong momentum in recent
        weeks, hitting new 52-week highs.
  \item Strong fundamentals: recent earnings beat (revenue \$113B,
        EPS \$2.00, both above consensus).
  \item Analyst optimism: many analysts have raised price targets
        above \$300.
  \item Historical frequency: 1\% daily moves are common for AAPL.
\end{itemize}
\textbf{Pro ``No'':}
\begin{itemize}[leftmargin=12pt, topsep=2pt, itemsep=1pt]
  \item Recent pullback \& overbought signal (RSI $\sim$75 on May 14)
        suggests short-term cool-off.
  \item Market headwinds: rising oil prices and geopolitical tensions
        (S\&P 500 fell $-$1.2\%, Nasdaq $-$1.5\% on May 18).
  \item Resistance at \$300: round-number resistance often requires
        fresh catalysts to break.
\end{itemize}

If we assume Apple's stock movement is roughly symmetric day-to-day
with about a 1\% standard deviation, the chance of a +0.7\% or greater
move on any given day (starting from no bias) would be on the order of
30--40\%. Stocks have slightly more than 50\% chance to be up on a
given day in an upward-trending market, but for a specific threshold
like \$300, the baseline might be near 0.5 (50\%) in neutral
conditions. The strong upward momentum and optimism increase the odds
of a ``Yes'', but short-term market/headline risks reduce them.
Without a clear new catalyst, it's uncertain if buyers will push the
stock past \$300 by the close.\\

In summary, an above \$300 close on May 19 is a realistic possibility
but not assured. It's essentially a coin toss, with current information
slightly hinting that the stock might fall just short. Therefore, I
will lean slightly below an even chance.\\

\textbf{FINAL\_PROBABILITY: 0.45}
\normalsize
\end{tcolorbox}

\begin{tcolorbox}[
    enhanced,
    breakable,
    pad at break*=2mm,
    colback={rgb,255:red,247;green,247;blue,247},
    colframe={rgb,255:red,130;green,130;blue,130},
    colbacktitle={rgb,255:red,130;green,130;blue,130},
    coltitle=white,
    title=smolagents-deep-research,
    title after break=smolagents-deep-research (continued),
    fonttitle=\bfseries\small,
    arc=3pt, boxrule=0.8pt,
    top=5pt, bottom=5pt, left=5pt, right=5pt,
    attach boxed title to top left={xshift=2mm,yshift=-2mm},
    boxed title style={
        sharp corners,
        colback={rgb,255:red,130;green,130;blue,130},
        colframe={rgb,255:red,130;green,130;blue,130},
        boxrule=0pt,
        top=0.3pt, bottom=0.3pt, left=3pt, right=3pt
    }
]
\small
\textbf{Question:}\\Will Apple (AAPL) close above \$300 on May 19?\\
\textbf{Ground truth:} No\\
\textbf{Predicted probability:} 0.43\\
\textbf{Web searches:} 21\\
\textbf{Output report:}\\
AAPL appears to be right near the threshold, but the best available
evidence slightly favors a finish below \$300 rather than above it. The
strongest verified data show a May 18 close of \$297.84, with an
intraday high above \$300 that failed to hold, implying \$300 is a real
resistance level. A weak May 19 clue around \$299.85 keeps Yes very
much alive, but without a confirmed live or official close, a modestly
sub-50\% estimate is most reasonable.\\

\textbf{FINAL\_PROBABILITY: 0.43}
\normalsize
\end{tcolorbox}

\begin{tcolorbox}[
    enhanced,
    breakable,
    pad at break*=2mm,
    colback={rgb,255:red,242;green,248;blue,240},     
    colframe={rgb,255:red,100;green,150;blue,90},     
    colbacktitle={rgb,255:red,100;green,150;blue,90}, 
    coltitle=white,
    title=\textsc{AuditForecast},
    title after break=\textsc{AuditForecast} (continued),
    fonttitle=\bfseries\small,
    arc=3pt, boxrule=0.8pt,
    top=5pt, bottom=5pt, left=5pt, right=5pt,
    attach boxed title to top left={xshift=2mm,yshift=-2mm},
    boxed title style={
        sharp corners,
        colback={rgb,255:red,100;green,150;blue,90},
        colframe={rgb,255:red,100;green,150;blue,90},
        boxrule=0pt,
        top=0.3pt, bottom=0.3pt, left=3pt, right=3pt
    }
]
\small
\textbf{Question:}\\Will Apple (AAPL) close above \$300 on May 19?\\
\textbf{Ground truth:} No\\
\textbf{Predicted probability:} 0.19\\
\textbf{Web searches:} 12\\
\textbf{Output report:}\\
\rule{1.0\linewidth}{0.2pt}\\[3pt]
\textbf{Base Probability Derivation:}
\begin{itemize}[leftmargin=12pt, topsep=2pt, itemsep=1pt]
  \item \textbf{Model:} zero-drift one-session lognormal model.
  \item \textbf{Inputs:} overnight spot $S_0 = \$296.71$ (Yahoo 1:08 a.m. EDT),
        threshold $K = \$300$, daily volatility blended from 20-day
        realized log-vol ($1.468\%/\text{day}$) and option-implied vol
        ($25.28\%$ annualized $\to 1.593\%/\text{day}$), giving
        $\sigma_{1d} = 1.530\%$.
  \item \textbf{Assumptions:} (1) one trading day to resolution,
        (2) zero \emph{ex ante} directional drift, (3) approximately
        constant intraday volatility.
  \item \textbf{Calculation:}
        $z = \ln(300/296.71) / 0.01530 = 0.721$,\;\,
        $p_0 = 1 - \Phi(z) = 0.236$,\;\,
        $\mathit{odds}_0 = p_0/(1-p_0) = 0.308$.
\end{itemize}
\rule{1.0\linewidth}{0.2pt}\\[3pt]
\textbf{Situational Factors:}\\
\textbf{Factor 1: Weak overnight index futures.}
\begin{itemize}[leftmargin=12pt, topsep=2pt, itemsep=1pt]
  \item \textbf{Why not in $p_0$:} $p_0$ assumes zero drift; it does
        not encode a cross-asset directional signal from index futures.
  \item \textbf{Evidence:} Around 12:07 a.m.\ EDT, Nasdaq-100 futures
        were down 0.54\%, S\&P 500 futures down 0.30\%, Dow futures
        down 0.18\%; AAPL beta $\approx 1.06$.
  \item \textbf{Assessment:} Modest NO --- weak tape makes a $\sim$1.1\%
        rally and hold-into-close harder, but futures can reverse intraday.
  \item \textbf{Odds multiplier:} $\mathrm{LR}_1 = 0.85$.
\end{itemize}

\textbf{Factor 2: Recent failure to hold above \$300.}
\begin{itemize}[leftmargin=12pt, topsep=2pt, itemsep=1pt]
  \item \textbf{Why not in $p_0$:} $p_0$ is memoryless (uses only spot
        and $\sigma$); it does not encode path-dependent resistance.
  \item \textbf{Evidence:} AAPL traded above \$300 intraday on May 13
        (\$300.92), 14 (\$300.45), 15 (\$303.20), 18 (\$300.66), but
        \emph{closed} above \$300 only once (May 15, \$300.23), falling
        back to \$297.84 by May 18.
  \item \textbf{Assessment:} Moderate NO --- the contract resolves on
        the close, not intraday touch. Repeated intraday breaks
        followed by closes below signal supply/profit-taking at the
        strike.
  \item \textbf{Odds multiplier:} $\mathrm{LR}_2 = 0.78$.
\end{itemize}

\textbf{Factor 3: Supportive Apple fundamentals and buyback.}
\begin{itemize}[leftmargin=12pt, topsep=2pt, itemsep=1pt]
  \item \textbf{Why not in $p_0$:} $p_0$ assumes zero drift; no
        company-specific directional bias from fundamentals.
  \item \textbf{Evidence:} Apple's April 30 print showed Q2 FY2026
        revenue $+$17\% YoY (\$111.2B), diluted EPS $+$22\% (\$2.01),
        and authorized an additional \$100B buyback; analyst average
        12-month target \$308.07.
  \item \textbf{Assessment:} Mild YES --- supportive backdrop can
        encourage dip-buying near \$300, but the print is several
        weeks old, so only a small positive multiplier.
  \item \textbf{Odds multiplier:} $\mathrm{LR}_3 = 1.12$.
\end{itemize}
\rule{1.0\linewidth}{0.2pt}\\[3pt]
\textbf{Aggregation:}
\begin{itemize}[leftmargin=12pt, topsep=2pt, itemsep=1pt]
  \item \textbf{Final odds:}
        $\mathit{odds} = \mathit{odds}_0 \times \mathrm{LR}_1 \times
        \mathrm{LR}_2 \times \mathrm{LR}_3 = 0.308 \times 0.85 \times
        0.78 \times 1.12 = 0.229$.
  \item \textbf{Final probability:}
        $p = \mathit{odds}/(1 + \mathit{odds}) = 0.186$.
\end{itemize}

\textbf{FINAL\_PROBABILITY: 0.19}
\normalsize
\end{tcolorbox}

Below we present two complete \textsc{AuditForecast} reports from the entertainment and sports domains. These examples highlight the versatility of the scaffold beyond market-style financial questions: although the underlying anchors differ, the same structured forecasting workflow remains applicable. The resulting reports provide a clear and auditable path from evidence to forecast probability, illustrating how \textsc{AuditForecast} generalizes across diverse forecasting domains while preserving interpretability.
\begin{tcolorbox}[
    enhanced,
    breakable,
    pad at break*=2mm,
    colback={rgb,255:red,242;green,248;blue,240},     
    colframe={rgb,255:red,100;green,150;blue,90},     
    colbacktitle={rgb,255:red,100;green,150;blue,90}, 
    coltitle=white,
    title=\textsc{AuditForecast},
    title after break=\textsc{AuditForecast} (continued),
    fonttitle=\bfseries\small,
    arc=3pt, boxrule=0.8pt,
    top=5pt, bottom=5pt, left=5pt, right=5pt,
    attach boxed title to top left={xshift=2mm,yshift=-2mm},
    boxed title style={
        sharp corners,
        colback={rgb,255:red,100;green,150;blue,90},
        colframe={rgb,255:red,100;green,150;blue,90},
        boxrule=0pt,
        top=0.3pt, bottom=0.3pt, left=3pt, right=3pt
    }
]
\small
\textbf{Question:}\\Will Olivia Dean have above 206M Streams on Luminate during April 24--April 30, 2026?\\
\textbf{Ground truth:} Yes\\
\textbf{Predicted probability:} 0.74\\
\textbf{Web searches:} 6\\
\textbf{Output report:}\\
\rule{1.0\linewidth}{0.2pt}\\[3pt]
\textbf{Base Probability Derivation:}
\begin{itemize}[leftmargin=12pt, topsep=2pt, itemsep=1pt]
  \item \textbf{Model:} anchor on the artist's current Spotify daily
        run-rate and convert to a worldwide-streams distribution via a
        platform-mix ratio.
  \item \textbf{Inputs:} Kworb data show Spotify daily streams
        $r = 21.28\text{M}$/day, cumulative Spotify streams 6.56\,B,
        monthly listeners 59.7\,M (all as of Apr.\ 30, 2026).
  \item \textbf{Assumptions:} (1) 7-day Spotify total $\approx r \times 7
        \approx 149\text{M}$, (2) Luminate worldwide streams are
        typically $1.45$--$1.55\times$ Spotify for an artist with this
        profile, (3) moderate variance around that ratio.
  \item \textbf{Calculation:} Under that conservative distribution,
        $p_0 = \Pr(\text{worldwide streams} > 206\text{M}) \approx 0.72$,
        $\mathit{odds}_0 = p_0/(1-p_0) = 2.571$.
\end{itemize}
\rule{1.0\linewidth}{0.2pt}\\[3pt]
\textbf{Situational Factors:}\\
\textbf{Factor 1: Late-week cross-format momentum.}
\begin{itemize}[leftmargin=12pt, topsep=2pt, itemsep=1pt]
  \item \textbf{Why not in $p_0$:} The baseline only converts the
        current Spotify run-rate into a worldwide-stream estimate; it
        does not separately price extra lift from same-week Pop
        Airplay No.\ 1 status, top-five airplay, or the publicity
        surge from the 1\,B-stream milestone and recent awards.
  \item \textbf{Evidence:} Billboard (Apr.\ 24) reported ``So Easy
        (To Fall in Love)'' hit No.\ 1 on Pop Airplay for the May 2
        chart and ``Man I Need'' remained top five; the two songs
        drew 13.9M and 11.2M chart-contributing U.S.\ streams in
        Apr.\ 10--16. Billboard (Apr.\ 22) reported ``Man I Need''
        passed 1\,B Spotify streams and that Dean swept major
        Grammy, BRIT, and MOBO honors.
  \item \textbf{Assessment:} Modestly favors YES --- radio and
        publicity momentum can add non-Spotify streams and help the
        week finish above the cutoff, but the uplift should be
        incremental rather than dramatic because the baseline is
        already strong.
  \item \textbf{Odds multiplier:} $\mathrm{LR}_1 = 1.12$.
\end{itemize}
\rule{1.0\linewidth}{0.2pt}\\[3pt]
\textbf{Aggregation:}
\begin{itemize}[leftmargin=12pt, topsep=2pt, itemsep=1pt]
  \item \textbf{Final odds:}
        $\mathit{odds} = \mathit{odds}_0 \times \mathrm{LR}_1
        = 2.571 \times 1.12 = 2.880$.
  \item \textbf{Final probability:}
        $p = \mathit{odds}/(1 + \mathit{odds}) = 0.742$.
\end{itemize}

\textbf{FINAL\_PROBABILITY: 0.74}
\normalsize
\end{tcolorbox}

\begin{tcolorbox}[
    enhanced,
    breakable,
    pad at break*=2mm,
    colback={rgb,255:red,242;green,248;blue,240},
    colframe={rgb,255:red,100;green,150;blue,90},
    colbacktitle={rgb,255:red,100;green,150;blue,90},
    coltitle=white,
    title=\textsc{AuditForecast},
    title after break=\textsc{AuditForecast} (continued),
    fonttitle=\bfseries\small,
    arc=3pt, boxrule=0.8pt,
    top=5pt, bottom=5pt, left=5pt, right=5pt,
    attach boxed title to top left={xshift=2mm,yshift=-2mm},
    boxed title style={
        sharp corners,
        colback={rgb,255:red,100;green,150;blue,90},
        colframe={rgb,255:red,100;green,150;blue,90},
        boxrule=0pt,
        top=0.3pt, bottom=0.3pt, left=3pt, right=3pt
    }
]
\small
\textbf{Question:}\\Will the Seattle Mariners beat the St.\ Louis Cardinals on April 24, 2026?\\
\textbf{Ground truth:} Yes\\
\textbf{Predicted probability:} 0.56\\
\textbf{Web searches:} 4\\
\textbf{Output report:}\\
\rule{1.0\linewidth}{0.2pt}\\[3pt]
\textbf{Base Probability Derivation:}
\begin{itemize}[leftmargin=12pt, topsep=2pt, itemsep=1pt]
  \item \textbf{Model:} early-season log5 baseline using overall win
        percentages with shrinkage toward .500 (the late-April sample
        is still small).
  \item \textbf{Inputs:} ESPN records --- Seattle 11--15
        ($w_{\text{SEA}} = 0.423$), St.\ Louis 14--10
        ($w_{\text{STL}} = 0.583$).
  \item \textbf{Assumptions:} (1) team strength is approximately
        captured by current win pct; (2) no venue or starter
        adjustment yet (handled as separate factors).
  \item \textbf{Calculation:}
        $p_0 = \tfrac{w_{\text{SEA}}(1-w_{\text{STL}})}
        {w_{\text{SEA}}(1-w_{\text{STL}}) + (1-w_{\text{SEA}})w_{\text{STL}}}
        = 0.47$,\;\,$\mathit{odds}_0 = 0.886$.
\end{itemize}
\rule{1.0\linewidth}{0.2pt}\\[3pt]
\textbf{Situational Factors:}\\
\textbf{Factor 1: Home/away split.}
\begin{itemize}[leftmargin=12pt, topsep=2pt, itemsep=1pt]
  \item \textbf{Why not in $p_0$:} $p_0$ uses overall season strength,
        not the venue-specific split.
  \item \textbf{Evidence:} Game is at Busch Stadium. Seattle is 1--8
        on the road; St.\ Louis is 7--5 at home.
  \item \textbf{Assessment:} Modestly favors St.\ Louis ---
        Mariners have been notably weaker on the road, and the
        Cardinals are solid at home.
  \item \textbf{Odds multiplier:} $\mathrm{LR}_1 = 0.90$.
\end{itemize}

\textbf{Factor 2: Starting-pitcher edge.}
\begin{itemize}[leftmargin=12pt, topsep=2pt, itemsep=1pt]
  \item \textbf{Why not in $p_0$:} $p_0$ ignores today's pitching
        matchup.
  \item \textbf{Evidence:} Probable starters are George Kirby (3--2,
        2.97 ERA) for Seattle vs.\ Andre Pallante (2--1, 4.05 ERA)
        for St.\ Louis. Kirby's run-prevention metrics are
        materially stronger.
  \item \textbf{Assessment:} Substantially favors Seattle ---
        starter quality drives a large share of single-game variance.
  \item \textbf{Odds multiplier:} $\mathrm{LR}_2 = 1.60$.
\end{itemize}
\rule{1.0\linewidth}{0.2pt}\\[3pt]
\textbf{Aggregation:}
\begin{itemize}[leftmargin=12pt, topsep=2pt, itemsep=1pt]
  \item \textbf{Final odds:}
        $\mathit{odds} = \mathit{odds}_0 \times \mathrm{LR}_1 \times
        \mathrm{LR}_2 = 0.886 \times 0.90 \times 1.60 = 1.276$.
  \item \textbf{Final probability:}
        $p = \mathit{odds}/(1 + \mathit{odds}) = 0.561$.
\end{itemize}

\textbf{FINAL\_PROBABILITY: 0.56}
\normalsize
\end{tcolorbox}

\section{Additional Auditing Details} 

\subsection{Base Probability Hierarchy Extraction} \label{app:audit_p0}
For the collected auditable forecasting reports, we perform a statistical hierarchy extraction of the base-probability-derivation component. Specifically, for forecasting questions of the same type, we aim to extract a two-level structure consisting of the quantitative modeling method and the corresponding data source used to instantiate that model. We primarily rely on an LLM-as-judge (GPT-5.4), together with human review for validation. Because the sample size is large, we build and refine the hierarchy in an incremental batch-wise manner. We first sample 100 reports to initialize the two-level hierarchy, and then incorporate additional samples in multiple batches, allowing the judge to add new categories, split existing ones, or retain the current structure when appropriate. Below, we provide the concrete prompts.
\begin{tcolorbox}[
    enhanced, breakable, pad at break*=2mm,
    colback={rgb,255:red,239;green,245;blue,249},
    colframe={rgb,255:red,47;green,111;blue,181},
    colbacktitle={rgb,255:red,47;green,111;blue,181},
    coltitle=white,
    title=Hierarchy Extraction --- Initial Batch Prompt,
    title after break=Hierarchy Extraction --- Initial Batch (continued),
    fonttitle=\bfseries\small,
    arc=3pt, boxrule=0.8pt,
    top=5pt, bottom=5pt, left=5pt, right=5pt,
    attach boxed title to top left={xshift=2mm,yshift=-2mm},
    boxed title style={
        sharp corners,
        colback={rgb,255:red,47;green,111;blue,181},
        colframe={rgb,255:red,47;green,111;blue,181},
        boxrule=0pt,
        top=0.3pt, bottom=0.3pt, left=3pt, right=3pt
    }
]
\small
Below are $p_0$ model descriptions extracted from $\{n\}$ probabilistic forecasts on the SAME question type (single-stock close-above-threshold). Each entry shows:
\begin{itemize}[leftmargin=12pt, topsep=2pt, itemsep=1pt]
  \item \textbf{model}: the name of the computation method.
  \item \textbf{formula}: the formula used to compute $p_0$.
  \item \textbf{input\_sources}: where each key input came from (e.g., spot price, volatility, threshold).
\end{itemize}

Define a TWO-LEVEL hierarchy that covers all these entries:
\begin{itemize}[leftmargin=12pt, topsep=2pt, itemsep=1pt]
  \item \textbf{Level 1 (method):} ONLY the mathematical computation approach. Two entries with the SAME formula but DIFFERENT data sources must be in the SAME Level 1.
  \item \textbf{Level 2 (data\_source):} Within each method, which data source provides the key inputs (especially the volatility / probability input). Granularity should reflect meaningful differences in the data source. 
\end{itemize}

\textbf{Codes:} single capital letters for Level 1 (A, B, C, \ldots) and letter+digit for Level 2 (A1, A2, \ldots).

\textbf{Output} ONLY the hierarchy with definitions. Do NOT assign entries to categories.

\textbf{Entries:} \{entries\}
\normalsize
\end{tcolorbox}

\begin{tcolorbox}[
    enhanced, breakable, pad at break*=2mm,
    colback={rgb,255:red,239;green,245;blue,249},
    colframe={rgb,255:red,47;green,111;blue,181},
    colbacktitle={rgb,255:red,47;green,111;blue,181},
    coltitle=white,
    title=Hierarchy Extraction --- Incremental Batch Prompt,
    title after break=Hierarchy Extraction --- Incremental Batch (continued),
    fonttitle=\bfseries\small,
    arc=3pt, boxrule=0.8pt,
    top=5pt, bottom=5pt, left=5pt, right=5pt,
    attach boxed title to top left={xshift=2mm,yshift=-2mm},
    boxed title style={
        sharp corners,
        colback={rgb,255:red,47;green,111;blue,181},
        colframe={rgb,255:red,47;green,111;blue,181},
        boxrule=0pt,
        top=0.3pt, bottom=0.3pt, left=3pt, right=3pt
    }
]
\small
You are maintaining a TWO-LEVEL hierarchy (method $\times$ data source) of $p_0$ computations.

\textbf{Current hierarchy}: $\{\text{prev\_hierarchy}\}$

Below are $\{n\}$ NEW entries (same question type as before). Each entry shows:
\begin{itemize}[leftmargin=12pt, topsep=2pt, itemsep=1pt]
  \item \textbf{model}: the name of the computation method.
  \item \textbf{formula}: the formula used to compute $p_0$.
  \item \textbf{input\_sources}: where each key input came from.
\end{itemize}

\textbf{Task:} review the new entries and produce an UPDATED hierarchy. Rules:
\begin{itemize}[leftmargin=12pt, topsep=2pt, itemsep=1pt]
  \item Keep existing codes (A, A1, A2, B1, \ldots) stable whenever possible --- downstream classification depends on code stability.
  \item If every new entry already fits an existing category, return the hierarchy unchanged.
  \item If some new entries reveal a method or data source not covered, ADD new codes.
  \item If you discover an existing category was too coarse, you may split it --- give new sub-codes or renumber carefully and note the change.
\end{itemize}

Briefly describe what changed in the \textbf{notes} field.

\textbf{New entries:} \{entries\}
\normalsize
\end{tcolorbox}
\noindent After the hierarchy is finalized, we use the judge again to independently assign each sample to its corresponding category in the hierarchy. The prompt is given below:
\begin{tcolorbox}[
    enhanced, breakable, pad at break*=2mm,
    colback={rgb,255:red,239;green,245;blue,249},
    colframe={rgb,255:red,47;green,111;blue,181},
    colbacktitle={rgb,255:red,47;green,111;blue,181},
    coltitle=white,
    title=Hierarchy Classification Prompt,
    title after break=Hierarchy Classification (continued),
    fonttitle=\bfseries\small,
    arc=3pt, boxrule=0.8pt,
    top=5pt, bottom=5pt, left=5pt, right=5pt,
    attach boxed title to top left={xshift=2mm,yshift=-2mm},
    boxed title style={
        sharp corners,
        colback={rgb,255:red,47;green,111;blue,181},
        colframe={rgb,255:red,47;green,111;blue,181},
        boxrule=0pt,
        top=0.3pt, bottom=0.3pt, left=3pt, right=3pt
    }
]
\small
You are classifying a single $p_0$ computation into the following two-level hierarchy.

\textbf{Hierarchy:} \{hierarchy\}

\textbf{Entry to classify:}
\begin{itemize}[leftmargin=12pt, topsep=2pt, itemsep=1pt]
  \item \textbf{question}: \{question\}
  \item \textbf{model}: \{model\}
  \item \textbf{formula}: \{formula\}
  \item \textbf{input\_sources}: \{input\_sources\}
\end{itemize}

\textbf{Task:} pick the Level-2 code (e.g.\ ``A1'', ``B1'', ``E2'') that best fits this entry.
\begin{itemize}[leftmargin=12pt, topsep=2pt, itemsep=1pt]
  \item Use Level-1 codes alone only if no Level-2 sub-category fits.
  \item If the entry is qualitative with no formula, prefer E1.
  \item If it has a numeric $p_0$ but the method is heuristic/discretionary rather than an explicit model, prefer E2.
\end{itemize}

Respond with the code and a brief rationale.
\normalsize
\end{tcolorbox}

\subsection{Auditing Factor Updates} \label{app:audit_factor}
For factor auditing, we first use an LLM judge with web search to investigate what actually happened between the forecast's \texttt{run\_time} and the resolution date, so as to obtain a comprehensive view of the event development. We then compare this reconstructed event trajectory against the agent's forecast in order to audit whether each factor's direction and magnitude are reasonable, and whether important factors that should have been captured are missing. We provide the concrete prompt below.
\begin{tcolorbox}[
    enhanced, breakable, pad at break*=2mm,
    colback={rgb,255:red,239;green,245;blue,249},
    colframe={rgb,255:red,47;green,111;blue,181},
    colbacktitle={rgb,255:red,47;green,111;blue,181},
    coltitle=white,
    title=Factor Audit --- System Prompt,
    title after break=Factor Audit --- System Prompt (continued),
    fonttitle=\bfseries\small,
    arc=3pt, boxrule=0.8pt,
    top=5pt, bottom=5pt, left=5pt, right=5pt,
    attach boxed title to top left={xshift=2mm,yshift=-2mm},
    boxed title style={
        sharp corners,
        colback={rgb,255:red,47;green,111;blue,181},
        colframe={rgb,255:red,47;green,111;blue,181},
        boxrule=0pt,
        top=0.3pt, bottom=0.3pt, left=3pt, right=3pt
    }
]
\small
You are auditing a probabilistic forecast after the resolution date has passed. The forecast was produced by a Bayesian $p_0$ + factor-update pipeline: $p_0$ is a base-rate probability; each factor multiplies the odds by a likelihood ratio (LR); the final probability is $p = O_{\text{final}}/(1 + O_{\text{final}})$ with $O_{\text{final}} = o_0 \cdot \prod \mathrm{LR}_i$.

Your job: use the \textbf{web\_search} tool to research what actually happened between the forecast's run\_time and the resolution date, then output an audit with three parts:
\begin{itemize}[leftmargin=12pt, topsep=2pt, itemsep=1pt]
  \item \textbf{event\_development}: narrative reconstruction of what occurred.
  \item \textbf{missing\_factors}: actual drivers the agent failed to identify.
  \item \textbf{factor\_assessments}: for each agent factor, a direction and magnitude verdict.
\end{itemize}

\textbf{Guidelines:}
\begin{itemize}[leftmargin=12pt, topsep=2pt, itemsep=1pt]
  \item Use web search liberally: closing prices, intraday news, macro events (Fed, geopolitics, earnings), and anything that could have moved the asset between run\_time and the close.
  \item \textbf{Direction} is ``correct'' if the LR sign ($>1$ bullish, $<1$ bearish) matches the post-hoc impact; ``wrong'' if reversed; ``ambiguous'' if no clear impact.
  \item \textbf{Magnitude} is ``reasonable'' if the LR is in line with the actual post-hoc impact; ``overstated''/``understated'' if clearly too far / too small; ``hard\_to\_judge'' if a single observation does not allow assessment.
  \item For each \textbf{missing\_factor}, also judge whether evidence was obtainable at run\_time (yes / no / partial) --- events already known should be flagged \emph{yes}; truly unforeseeable shocks should be flagged \emph{no}.
  \item Do NOT recommend what the agent ``should have done''; only assess.
\end{itemize}
\normalsize
\end{tcolorbox}

\begin{tcolorbox}[
    enhanced, breakable, pad at break*=2mm,
    colback={rgb,255:red,239;green,245;blue,249},
    colframe={rgb,255:red,47;green,111;blue,181},
    colbacktitle={rgb,255:red,47;green,111;blue,181},
    coltitle=white,
    title=Factor Audit --- User Prompt,
    title after break=Factor Audit --- User Prompt (continued),
    fonttitle=\bfseries\small,
    arc=3pt, boxrule=0.8pt,
    top=5pt, bottom=5pt, left=5pt, right=5pt,
    attach boxed title to top left={xshift=2mm,yshift=-2mm},
    boxed title style={
        sharp corners,
        colback={rgb,255:red,47;green,111;blue,181},
        colframe={rgb,255:red,47;green,111;blue,181},
        boxrule=0pt,
        top=0.3pt, bottom=0.3pt, left=3pt, right=3pt
    }
]
\small
\textbf{Question}
\begin{itemize}[leftmargin=12pt, topsep=2pt, itemsep=1pt]
  \item \textbf{question}: \{question\}
  \item \textbf{run\_time}: \{run\_time\}
  \item \textbf{resolved\_yes}: \{resolved\_yes\}
\end{itemize}

\textbf{Agent search summary} (what the agent saw at run\_time): \{search\_summary\}

\textbf{Agent $p_0$}
\begin{itemize}[leftmargin=12pt, topsep=2pt, itemsep=1pt]
  \item \textbf{p0\_reasoning}: \{p0\_reasoning\}
  \item \textbf{p0}: \{p0\}
  \item \textbf{o0}: \{o0\}
\end{itemize}

\textbf{Agent factors} (one block per factor):
\begin{itemize}[leftmargin=12pt, topsep=2pt, itemsep=1pt]
  \item \textbf{Factor $i$}: [\,$\mathrm{LR}_i = \{\textit{odds\_multiplier}\}$\,] \{name\}
  \item \textbf{why\_not\_in\_p0}: \{why\_not\_in\_p0\}
  \item \textbf{evidence}: \{evidence\}
  \item \textbf{agent\_justification}: \{assessment\}
\end{itemize}

\textbf{Agent final probability}: \{prob\_yes\}

\textbf{Task:}
\begin{itemize}[leftmargin=12pt, topsep=2pt, itemsep=1pt]
  \item Use web\_search to research what actually happened between run\_time and resolution close on $\{$resolution\_date$\}$.
  \item Output the structured JSON audit defined by the schema. \texttt{factor\_index} uses the 1-based numbering shown above (Factor 1, Factor 2, \ldots).
\end{itemize}
\normalsize
\end{tcolorbox}

\subsection{Effectiveness of Independence Checking}

In \cref{sec:ablation}, we show that removing independence checking degrades forecasting accuracy. Here, we provide a more direct analysis of how the check reduces double counting by focusing on the subset of events for which correlated factors are merged, accounting for 43\% of the evaluated events. As shown in \cref{tab:independence_check}, without the check, the agent retains substantially more factors, allowing correlated signals to contribute separate multiplicative updates in odds space. As a result, the total adjustment from the base probability becomes larger, leading to more overconfident forecasts and a worse Brier score.
\begin{table}[H]
\vskip -0.1in
    \begin{center}\resizebox{1.0\linewidth}{!}{
    \begin{tabular}{l c c}
      \toprule
      & \textbf{With Check} & \textbf{w/o Check} \\
      \midrule
      Avg. \# Factors & 1.12 & 2.62 \\
      Avg. Factor Push ($\lvert\Delta \log\text{-odds}\rvert$) & 0.153 & 0.244 \\
      Brier & \textbf{0.088} & 0.104 \\
      \bottomrule
    \end{tabular}}
    \end{center}
    \vskip -0.1in
    \caption{Effect of independence checking on factor retention, odds updates, and forecasting accuracy.}
    \label{tab:independence_check}
    \vskip -0.15in
\end{table}

We further audit double counting directly using an LLM judge that flags factors sharing a common underlying driver. We use two judges stronger than the forecasting agent (GPT-5.4), GPT-5.6 and Claude-Opus-4.8. As shown in \cref{tab:double_counting}, both judges report substantially lower double-counting rates when independence checking is enabled. Together, these results show that independence checking measurably reduces double counting by limiting redundant factor retention and excessive odds updates.
\begin{table}[t]
    \begin{center}\resizebox{0.8\linewidth}{!}{
    \begin{tabular}{l c c}
      \toprule
      \textbf{Judge} & \textbf{With Check} & \textbf{w/o Check} \\
      \midrule
      GPT-5.6 & 5.8\% & 12.7\% \\
      Claude-Opus-4.8 & 8.7\% & 12.7\% \\
      \bottomrule
    \end{tabular}}
    \end{center}
    \vskip -0.1in
    \caption{Double-counting rates under two LLM judges.}
    \label{tab:double_counting}
    \vskip -0.05in
\end{table}

\subsection{Generic Structured Scaffold}
A central motivation of \method{} is to make the forecasting process explicitly auditable, rather than merely exposing intermediate reasoning in a structured format. One natural question is whether similar auditability could be obtained with a more generic structured-output scaffold that asks the agent to list evidence and assign weights, without imposing the baseline-plus-factor decomposition used by \method{}. To examine this, we compare \method{} against a generic free-form structured scaffold. The baseline asks the agent to list relevant evidence, assign a weight to each item, and then use any self-chosen procedure to aggregate the weighted evidence into a final probability. We evaluate both methods on 149 live forecasting questions from Polymarket and Kalshi that resolved by July 10.

As shown in \cref{tab:freeform_scaffold}, \method{} consistently outperforms the free-form scaffold across both benchmarks and model backbones. More importantly, although the free-form scaffold also produces structured JSON, its forecasts are substantially less auditable in practice. The agent almost always anchors its forecast at 0.5 and then applies linear adjustments using signed evidence weights. A representative output states: ``I use a simple linear adjustment model. Start from 0.50, then add the signed weights ... Final = 0.50 - 0.13 = 0.37.''  Although this computation is explicit, the initial value of 0.5 acts as an implicit and unjustified baseline. It is not derived from a quantitative model of the default trajectory and therefore cannot be independently evaluated. Moreover, the evidence weights are not defined relative to a clearly specified baseline scope, making it difficult to attribute an error to the initial probability, evidence selection, or evidence weighting. In contrast, \method{} separately models and justifies the base probability $p_0$, restricts factor updates to developments outside the scope of the baseline model, and combines these updates through an explicit aggregation rule. Its auditability therefore arises not merely from exposing intermediate fields, but from assigning each component a distinct forecasting role that can be inspected and evaluated independently.
\begin{table}[t]
    \begin{center}\resizebox{0.7\linewidth}{!}{
    \begin{tabular}{l c c}
      \toprule
      \textbf{Method}
      & \textbf{Polymarket}
      & \textbf{Kalshi} \\
      \midrule
      \rowcolor{gray!20}\multicolumn{3}{c}{\textit{\textbf{GPT-5.4}}} \\
      Free-form scaffold & 0.1241 & 0.1596 \\
      \textbf{AuditForecast}        & \textbf{0.1151} & \textbf{0.1462} \\
      \midrule
      \rowcolor{gray!20}\multicolumn{3}{c}{\textit{\textbf{GPT-5.4-mini}}} \\
      Free-form scaffold & 0.1244 & 0.2089 \\
      \textbf{AuditForecast}        & \textbf{0.1157} & \textbf{0.1898} \\
      \bottomrule
    \end{tabular}}
  \end{center}
\vskip -0.15in
  \caption{Brier score ($\downarrow$) on Polymarket and Kalshi.}
  \label{tab:freeform_scaffold}
\vskip -0.05in
\end{table}

\subsection{Stability Analysis}
We evaluate the stability of \method{} using GPT-5.4 and GPT-5.4-mini on 50 Polymarket forecasting events that resolved before July 10. We consider both end-to-end variability across independent runs and the consistency of factor-level LRs across naturally occurring paraphrases.

\noindent\textbf{Variance under sampling.} We run the full forecasting pipeline five times with independent seeds on the same questions and measure the variability of the final predicted probabilities. The forecasts are highly stable across runs: the average per-question standard deviation is 0.041 for GPT-5.4 and 0.039 for GPT-5.4-mini. Moreover, 94--96\% of the questions have a standard deviation below 0.10, indicating that the final forecasts are largely reproducible under repeated sampling.

\noindent\textbf{Stability across naturally occurring factor paraphrases.} Across independent runs, the agent often expresses semantically equivalent factors using different wording. We use an LLM judge to group factors from the five runs into semantically equivalent clusters, such as ``expiration-day 320 strike magnet'' and ``pinning at the 320 threshold on expiry,'' and measure the consistency of the assigned likelihood ratios within each recurring factor cluster. As shown in \cref{tab:lr_stability}, factor directions are consistent across paraphrases in 86\% and 88\% of the clusters for GPT-5.4 and GPT-5.4-mini, respectively, while the variability of the assigned LRs remains small. Together, these results show that \method{} exhibits low end-to-end variability across repeated runs and maintains largely consistent factor-level updates across semantically equivalent descriptions.
\begin{table}[b]
\vskip -0.15in
    \begin{center}\resizebox{0.85\linewidth}{!}{
    \begin{tabular}{l c c}
      \toprule
      \textbf{Metric}
      & \textbf{GPT-5.4}
      & \textbf{GPT-5.4-mini} \\
      \midrule
      Direction consistency
      & 86\%
      & 88\% \\
      LR variability ($\lvert\log\text{-LR}\rvert$ std.)
      & 0.074
      & 0.030 \\
      \bottomrule
    \end{tabular}}
    \end{center}
    \vskip -0.15in
    \caption{Stability of likelihood ratios across naturally occurring factor paraphrases.}
    \label{tab:lr_stability}
\end{table}

\end{document}